%% file: main.tex
\documentclass[acmlarge]{acmart}

\AtBeginDocument{%
  \providecommand\BibTeX{{%
    \normalfont B\kern-0.5em{\scshape i\kern-0.25em b}\kern-0.8em\TeX}}}

\setcopyright{rightsretained}
\acmJournal{IMWUT}
\acmYear{2023} 
\acmVolume{7} 
\acmNumber{2} 
\acmArticle{75} 
\acmMonth{6} 
\acmPrice{}
\acmDOI{10.1145/3596234}

\usepackage{multirow}
\usepackage{subfiles}
\usepackage{subfig}
\usepackage{float}
\usepackage{ragged2e}
\usepackage{graphicx}
\usepackage{bigstrut}
\usepackage{algorithm2e}
\usepackage{cancel}

\newcommand{\edit}[1]{{\textcolor{black}{#1}}}

\begin{document}

\title{ConvBoost: Boosting ConvNets for Sensor-based Activity Recognition}

\author{Shuai Shao}
\affiliation{%
  \institution{Department of Computer Science, University of Warwick}
  \city{Coventry}
  \country{UK}}
\email{Shuai.Shao.1@warwick.ac.uk}
\orcid{0009-0002-7028-0944}

\author{Yu Guan}
\authornote{corresponding author}
\affiliation{%
  \institution{Department of Computer Science, University of Warwick}
  \city{Coventry}
  \country{UK}}
\email{Yu.Guan@warwick.ac.uk}
\orcid{0000-0002-1283-3806}

\author{Bing Zhai}
\affiliation{%
  \institution{Computer and Information Sciences, Northumbria University}
  \city{Newcastle upon Tyne}
  \country{UK}}
\email{bing.zhai@northumbria.ac.uk}
\orcid{0000-0003-1635-1406}

\author{Paolo Missier}
\affiliation{%
  \institution{School of Computing, Newcastle University}
  \city{Newcastle upon Tyne}
  \country{UK}}
\email{paolo.missier@newcastle.ac.uk}
\orcid{0000-0002-0978-2446}

\author{Thomas Pl{\"o}tz}
\affiliation{%
  \institution{School of Interactive Computing, Georgia Institute of Technology}
  \city{Atlanta}
  \country{USA}}
\email{thomas.ploetz@gatech.edu}
\orcid{0000-0002-1243-7563}

\renewcommand{\shortauthors}{Shao and Guan, et al.}
\begin{abstract}
\subfile{sections/abstract}
\end{abstract}

\ccsdesc[500]{Computing methodologies~Machine learning approaches}
\ccsdesc[500]{Human-centered computing~Ubiquitous and mobile computing}
	%
	%
	%
\keywords{Human Activity Recognition, Deep Learning, Ensemble, Data Augmentation, Sensors}

\maketitle

\section{Introduction}
\subfile{sections/introduction}

\section{Related Work}

\subfile{sections/related_work}

\section{The ConvBoost Framework}

\subfile{sections/method}

\section{Experimental Evaluation}
\subfile{sections/experiment}

\section{Conclusion}
\subfile{sections/conclusion}

\section*{Acknowledgments}
We would like to thank the anonymous reviewers for their constructive comments that helped improving the quality of this paper.

\bibliographystyle{ACM-Reference-Format}
\bibliography{main}


\appendix
\subfile{sections/appendix}

\end{document}

%% file: sections/abstract.tex
Human activity recognition (HAR) is one of the core research themes in ubiquitous and wearable computing.
With the shift to deep learning (DL) based analysis approaches, it has become possible to extract high-level features and perform classification in an end-to-end manner.
Despite their promising overall capabilities, DL-based HAR may suffer from overfitting due to the notoriously small, often inadequate, amounts of labeled sample data that are available for typical HAR applications. 
In response to such challenges, we propose ConvBoost -- a novel, three-layer, structured model architecture and boosting framework for convolutional network based HAR.
Our framework generates additional training data from three different perspectives for improved HAR, aiming to alleviate the shortness of labeled training data in the field.
Specifically, with the introduction of three conceptual layers--Sampling Layer, Data Augmentation Layer, and Resilient Layer--we develop three ``boosters''--R-Frame, Mix-up, and C-Drop--to enrich the per-epoch training data by dense-sampling, synthesizing, and simulating, respectively.
These new conceptual layers and boosters, that are universally applicable for any kind of convolutional network, have been designed based on the characteristics of the sensor data and the concept of frame-wise HAR.
In our experimental evaluation on three standard benchmarks (Opportunity, PAMAP2, GOTOV) we demonstrate the effectiveness of our ConvBoost framework for HAR applications based on variants of convolutional networks: vanilla CNN, ConvLSTM, and Attention Models.
We achieved substantial performance gains for all of them, which suggests that the proposed approach is generic and can serve as a practical solution for boosting the performance of existing ConvNet-based HAR models.
This is an open-source project, and the code can be found at https://github.com/sshao2013/ConvBoost

%% file: sections/introduction.tex
Human Activity Recognition (HAR) is a core research topic in ubiquitous and wearable computing. 
HAR research covers a variety of application scenarios, including but not limited to health assessments, sports tracking and coaching, sleep monitoring etc.\ \cite{plotz2012automatic, avci2010activity, hammerla2015pd, ladha2013climbax, bing2020, babystroke19}.
\edit{HAR models are essentially mapping functions,} which \edit{associate streams of sensor data to limited sets of} activity types. 
A traditional HAR pipeline includes sliding window segmentation, feature engineering or extraction, followed by developing pattern recognition or machine learning models \edit{for the actual activity recognition step}. 
Out of them, feature engineering tends to be a trial-and-error process, which can be time-consuming. 
For different HAR tasks, the optimal features may vary from case to case, making feature designing process expensive and less scalable. 
Recently, deep learning (DL) came into popularity for HAR modelling, which can learn  high-level features and perform activity classification in an end-to-end manner.
In many cases, they tend to be more effective than traditional feature engineering methods \cite{plotz2018deep}.

For DL-based HAR, two major approaches exist: frame-wise and sample-wise processing \cite{plotz2018deep}.
Frame-wise methods are \edit{considered mainstream}, and such methods train \edit{networks} on  \edit{segmented frames (via sliding window)}, \edit{to then map} the signal frames to activities.
Sample-wise approaches, however, are normally trained using sequential models (e.g., Long Short Term Memory, LSTM \cite{hammerla2016deep, guan2017}), which can build the mapping relationship from each signal sample (i.e., at each timestamp) to activities.
Although sample-wise methods \edit{have shown to be successful with regards to}  handling the challenging non-repetitive (i.e., sequential) activities \cite{hammerla2016deep, guan2017}, the \edit{frame-wise processing using variants of convolutional networks--ConvNets--}
tends to be better at dealing with the more common repetitive activities such as walking, running and cycling, etc. 
Given the effectiveness and simplicity, frame-wise ConvNet-based methods are now considered  mainstream in the HAR field \cite{ordonez2016deep,hammerla2016deep,yaguchi2020human,meyer2021cnn,chen2020metier,bai2020adversarial,tang2021selfhar,munzner2017cnn,yang2015deep}.

Despite the promising performance of DL-based approaches, they \edit{often} suffer from overfitting problems, \edit{especially in application scenarios where there are only small amounts of labeled example data available for model training}.
\edit{Unfortunately, this is a rather common problem due to logistical and privacy related restrictions, which render data annotation  expensive  if not impossible at times}. 
To alleviate this overfitting issue, several research directions were explored including data augmentation \cite{li2020activitygan, wang2018sensorygans, zhang2017mixup}, self-supervised learning (SSL)\cite{ssl_HAR19, tang2021selfhar, Haresamudram_ssl, Haresamudram_iswc20}, or learning paradigm design (e.g., ensemble learning \cite{guan2017, tan2022human, sekiguchi2020ensemble}, self-paced curriculum learning\cite{self-paced_curriculum_learning}, etc.). 
For HAR, SSL came into popularity in recent years, which can take advantage of unlabeled data for activity representation learning, with improved performance in downstream HAR tasks \cite{Haresamudram_ssl}. For ensemble learning, in \cite{guan2017} an epoch-wise bagging scheme was proposed, based on which a number of epoch-wise LSTMs were generated and combined for sample-wise HAR tasks.
To inject diversity for better ensemble results, some hyper-parameters (e.g., window length, sampling point, batch size) were modelled as per-epoch variables.
This scheme led to very promising results in challenging HAR scenarios, yet the diversity injection mechanism was specifically designed for sample-wise LSTM and thus may limit its application to the mainstream frame-wise ConvNets in HAR.

Motivated by this epoch-wise bagging idea, in this work we aim to build a more generic ConvNet-Boosting
\edit{(ConvBoost)} framework for mainstream HAR ConvNets. 
Compared with \cite{guan2017}, which focused on ensemble learning, our ConvBoost \edit{defines--and solves--a per-epoch training data generation problem}, which \edit{renders our overall approach} more flexible for various DL models.
Compared with the popular SSL-based approaches \cite{Haresamudram_ssl}, which can learn representation from the unlabeled data, we argue the potentials of the original labeled training sequence haven't been fully exploited, and via our apporach we can generate high-quality training frames to boost the performance of ConvNets.
In our ConvBoost framework, we \edit{define} three conceptual layers, namely:
\textit{i)} Sampling Layer;
\textit{ii)}Data Augmentation Layer; and
\textit{iii)} Resilient Layer.
\edit{Their introduction into model training aims at generating per-epoch, diverse training examples from different perspectives to extend the available sample data. 
Within this three-layer structure, 
three boosters are integrated:
\emph{i)} Random Framing (R-Frame) booster;
\emph{ii)} Mix-up booster; and 
\emph{iii)} Channel Dropout (C-Drop) booster. 
These boosters enrich the per-epoch training examples by dense-sampling (via R-Frame),  synthesizing (via Mix-up), or simulating problematic signals (via C-Drop). 
With the per-epoch generated additional data, robust epoch-wise classifiers can be trained, which can be used individually (i.e., best base classifier, referred to as \textit{Single-Best} mode) or jointly (i.e., via  \textit{Ensemble} \cite{guan2017} mode) for different HAR scenarios.
}
In \autoref{ConvBoost}, we demonstrate our proposed 3-layer ConvBoost framework, and the contributions of this paper can be summarized as follows:

\begin{figure}[t]
  \centering
  \includegraphics[width=0.9\textwidth]{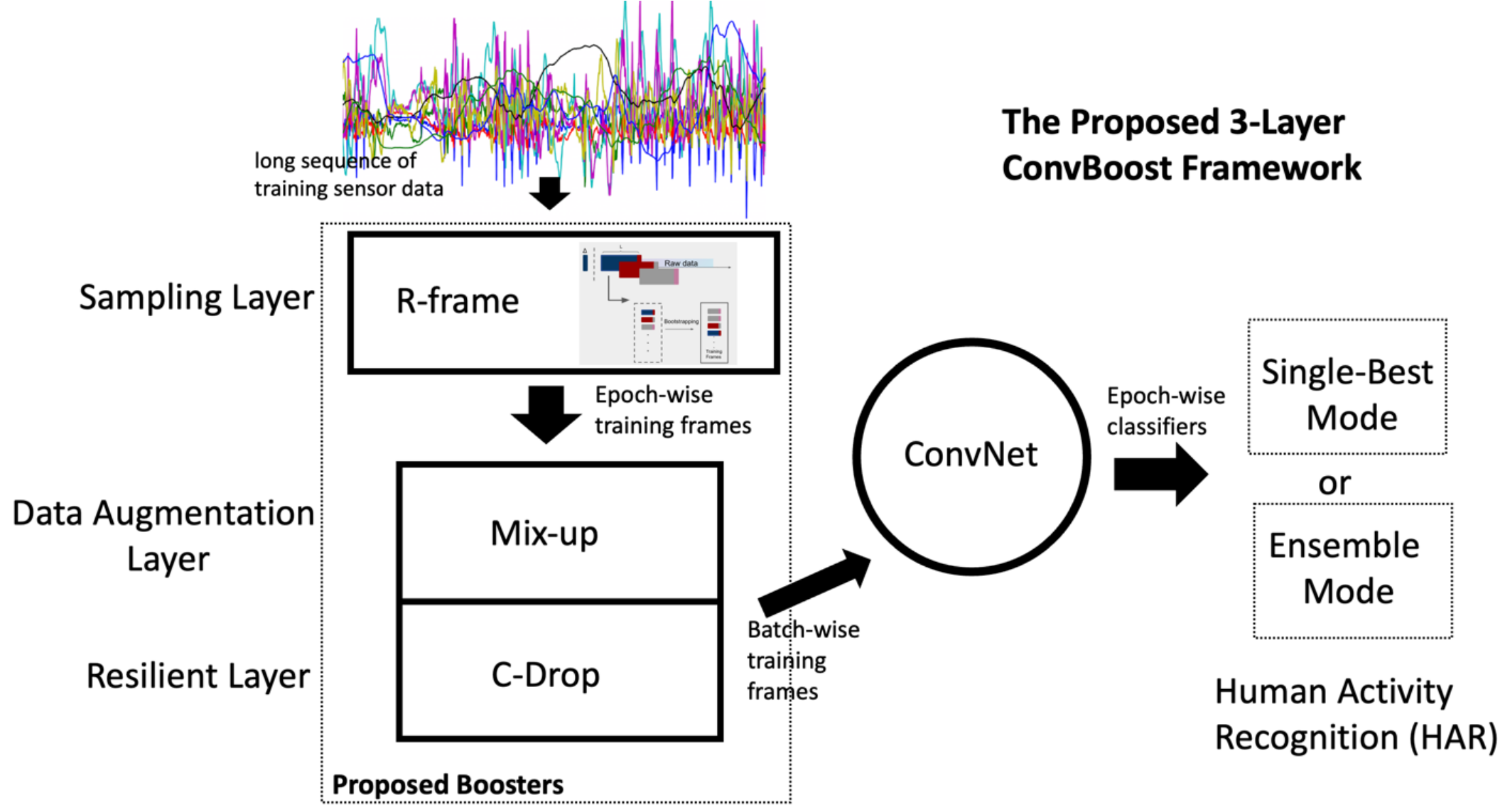}
  \caption{An overview of our proposed ConvBoost framework. Based on the three conceptual layers (and the corresponding boosters), additional per-epoch training frames can be generated to boost the performance of various types of ConvNets in HAR tasks.}
  \label{ConvBoost}
\end{figure}

\begin{enumerate}
\item 
\edit{We propose a 3-layer ConvBoost framework for mainstream ConvNet-based HAR. Through the three new conceptual layers, the proposed ConvBoost can dynamically generate per-epoch additional and complementary training examples for robust HAR model development. 
}

\item 
\edit{Based on the characteristics of sensor data and frame-based HAR problems, we design three tailored boosters corresponding to the three conceptual layers in our ConvBoost, which are Random Framing (R-Frame) booster, Mix-up booster, and Channel Dropout (C-Drop) booster. 
}

\item 
\edit{We demonstrate the effectiveness of our approach in an extensive experimental evaluation on three benchmark} datasets \edit{. Through our comprehensive experiments we gain an  understanding on the performance gains achieved by each layer/booster in our ConvBoost framework. }

\item 
Through our ConvBoost framework, we re-interpret the original epoch-wise bagging scheme \cite{guan2017}, and develop its ConvNet variant.
Through our experiments, we found that the performance gains of ConvBoost (Ensemble)  mainly stem from the strengthened classifiers (via the per-epoch additional complementary training data), instead of ensemble diversity, indicating training data generation can be a crucial direction for epoch-wise-ensemble based HAR.

\item 
\edit{Our ConvBoost is an extensible framework, and we explore two possible extensions: 
\emph{i)} extra boosters; and 
\emph{ii)} ConvBoost (Ensemble mode) compression.
}
\end{enumerate}

%% file: sections/related_work.tex
\subsection{ConvNets for Human Activity Recognition (HAR)}
Compared with traditional HAR, deep learning (DL) can extract high-level activity features and perform classification in an end-to-end manner.
With the promising performance, DL became mainstream in HAR research.   
Out of many DL architectures, the most popular ones are based on convolutional networks (ConvNets), e.g., CNN, ConvLSTM, and so on\cite{ronao2015deep, ordonez2016deep, hammerla2016deep, yang2015deep}. However, they may suffer from overfitting when with limited data or lack of annotations. 
Based on ConvNets, several research directions were explored to address this issue, including self-supervised learning (SSL) \cite{ssl_HAR19, tang2021selfhar, Haresamudram_ssl, Haresamudram_iswc20}, or data synthesis \cite{li2020activitygan, IMUtube}. 
The major aim of SSL is to learn representation from the unlabeled data by designing pre-training objectives or auxiliary tasks, before fine-tuning to the downstream HAR tasks. 
In \cite{ssl_HAR19} Saeed et al. developped a multi-task SSL scheme, and based on multiple auxiliary tasks (e.g., adding random noise, varying sampling rate), the activity representation can be learned and fine tuned with improved performance. 
In \cite{Haresamudram_iswc20}, a masked reconstruction based SSL was proposed, which demonstrated great improvements when compared with other unsupervised learning schemes in HAR tasks. 
In \cite{tang2021selfhar}, SelfHAR, a semi-supervised framework was proposed, which combined teacher-student self-training to exploit both unlabeled and labeled datasets while allowing for data augmentation, and multi-task self-supervision for improved activity representation learning. 
Haresamudram et al. comprehensively studied seven state-of-the-art SSL-based approaches on various HAR datasets \cite{Haresamudram_ssl}. 
Most recently, SSL was performed on a large-scale unlabeled activity dataset \cite{ssl_700000} for pre-trained model development, which demonstrated improved performance on other downstream HAR tasks.
Moreover, the popular generative models (e.g., GAN) was also explored in \cite{li2020activitygan} to generate synthetic activity sensor-data to tackle the lack of annotation problems.

\subsection{Epoch-wise Bagging Scheme for HAR}


To address the lack of annotation issue, Guan and Ploetz proposed a deep LSTM ensemble approach in \cite{guan2017} for better generalization. 
In their approach, the epoch-wise bagging scheme was proposed, which can simply generate per-epoch LSTM classifiers with nearly no extra training time cost.
The epoch-wise base classifiers were aggregated using a score-level fusion, which is simple and effective for challenging HAR tasks. 

However, directly applying the epoch-wise bagging scheme may face the lack of ensemble diversity problem since the base classifiers were generated across epochs using the same network structure.
One could use different network structures as base learners (with higher diversity), yet it would substantially increase the training costs. 
In \cite{guan2017}, it was pointed out that there would be no fusion effect if all the base learners were identical (i.e., zero diversity) and some diversity injection approaches were used to mitigate this problem in the epoch-wise bagging scheme.
Different from traditional deep learning approaches, in \cite{guan2017} some hyper-parameters of LSTM (i.e., initial sampling point, batch size, window/frame length) were modeled as random variables, whose values may vary across different epochs or batches. 
Although this hyper-parameter modelling strategy may inject some uncertainties for diverse epoch-wise classifier training, 
they were specifically designed for LSTM models (for sample-wise HAR), and cannot be applied directly to the mainstream ConvNet-based methods like CNN\cite{yang2015deep}, ConvLSTM\cite{morales2016deep}, Attention Model\cite{Murahari2018}, etc. 
Moreover, although LSTM ensemble \cite{guan2017} may significantly boost the performance (over the single LSTM), the diversity injection procedure is empirical and hasn't been studied using existing diversity measurement metrics (e.g., Q-Statistic (QS)\cite{udny1900association}).

In this section, we review previous HAR works for tackling the lack of annotation problem. 
For ConvNet-based approaches, SSL is one of the major research topics, aiming at learning representation via pre-train objectives or auxiliary tasks from unlabeled data to boost the performance of the downstream (supervised) HAR tasks. 
In this paper, we try to solve this problem from a different perspective. 
Instead of employing the unlabeled data, we argue the potentials of the original labeled sensor data haven't been fully exploited, and our ConvBoost framework can be used to generate high-quality labeled training data for improved HAR. 
It is worth noting that although the proposed ConvBoost framework is motivated by the epoch-wise bagging scheme \cite{guan2017}, its focus is per-epoch (labeled) training data generation, making it a very flexible and extensible solution.

%% file: sections/method.tex
\edit{In contrast to epoch-wise bagging scheme \cite{guan2017}, our ConvBoost framework is designed for mainstream HAR ConvNets, and the underlying mechanism is to create per-epoch training frames in three conceptual layers via dense-sampling, synthesizing, and simulating operations.
The training-frame generation nature 
}
makes it a flexible scheme that can be applicable to various ConvNet types.
\edit{The proposed three conceptual layers are Sampling Layer, Data Augmentation Layer, and Resilient Layer, and for each layer a booster is developed for per-epoch training frames generation, which are listed as follows: 
}

\begin{enumerate}
\item \emph{Sampling Layer}: in this layer we propose Random Framing (R-Frame) booster, which can be deemed as \edit{ an epoch-wise} dense-sampling approach, in contrast to traditional \edit{one-off} sliding window method. 

\item \emph{Data Augmentation Layer}: in this layer we apply mix-up booster, which can synthesize virtual data via interpolation \cite{zhang2017mixup}. 

\item \emph{Resilient Layer}: in this layer we randomly drop some sensor channels (i.e., Channel Dropout or C-Drop for short) to simulate the problematic sensor signals \edit{for diverse training frames.}
\end{enumerate}

Note these three layers are not limited to these three boosters (i.e., R-Frame, mix-up, and C-Drop), which can be used solely or jointly, or even replaced by other advanced boosters.
In the experimental section, we also study the performance gain achieved by booster combinations.

\subsection{Random Framing (R-Frame) Booster in Sampling Layer}
\edit{For many HAR systems, sliding window is a standard procedure, which converts long signal sequences into short, individual frames for classification.}  
\edit{For ConvNet-based HAR, these frames are normally shuffled before each training epoch.}  
However, DL-based approaches \edit{are notoriously data-hungry, and they often suffer from overfitting in scenarios where there is only limited amounts of labeled sample data available for model training}, which is a common problem in HAR.
\edit{
Although the per-epoch shuffling operation may reduce the overfitting effect to some extent, 
it is limited, since the overall training frames--constructed by the one-off sliding window--is a fixed set, which is problematic especially when training large ConvNets. 
}

\begin{figure}[t]
  \centering
\includegraphics[width=0.8\textwidth]{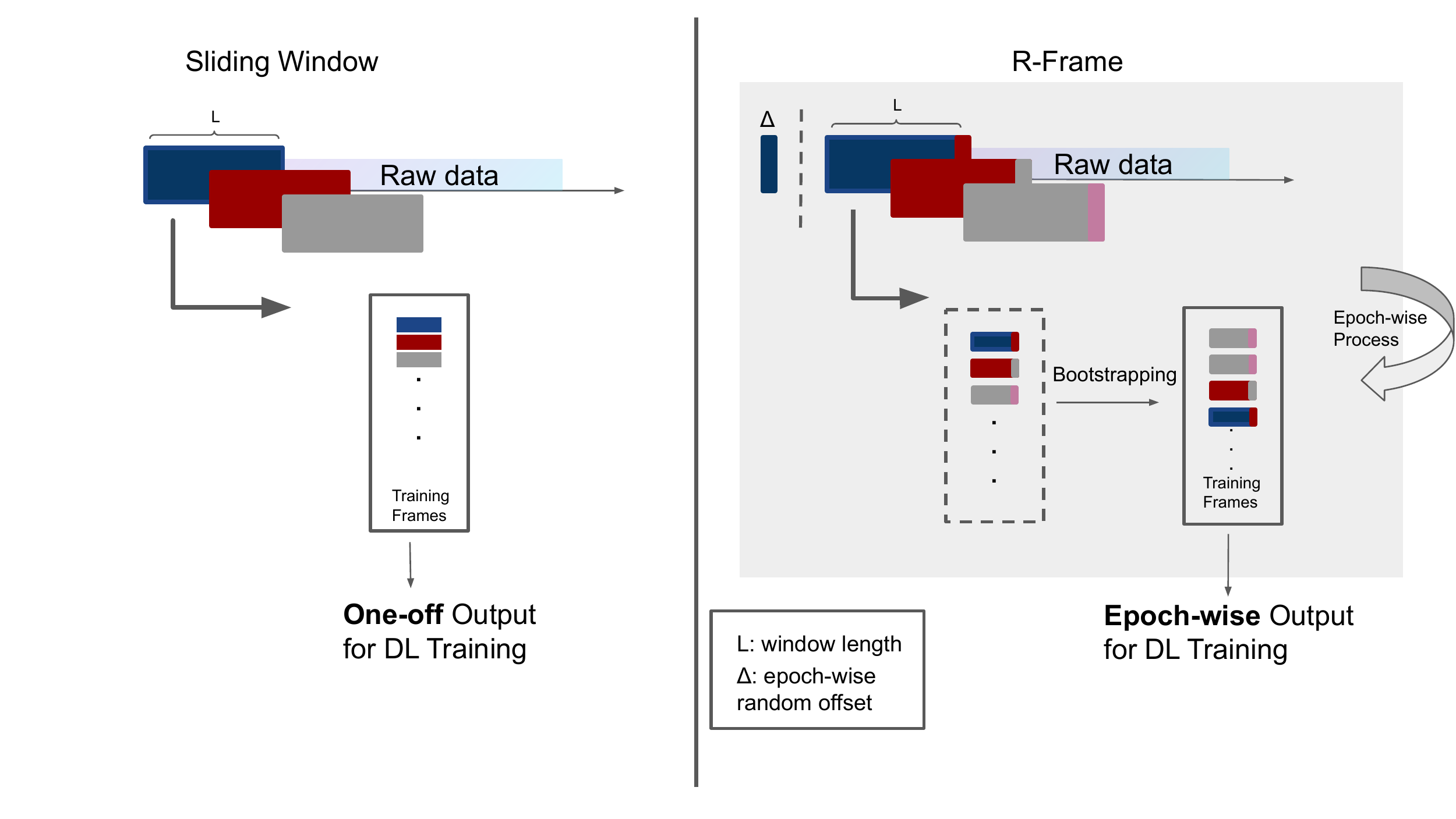}
    
  \caption{Traditional sliding window approach (left) and the proposed R-Frame booster (right) in the Sampling layer which can generate diverse per-epoch training frames for DL-based HAR}
  \label{r_frame}
\end{figure}

To address this issue, here we propose a 'Random Framing (R-Frame) booster' in the Sampling Layer, which can \edit{generate per-epoch \textit{dynamic} frame sets, 
in contrast to a \textit{fixed} frame set} produced by traditional sliding window. 
\edit{
In \autoref{r_frame}, we show both frame-generation approaches for ConvNets. 
Compared to traditional sliding window, we see:
}
\begin{itemize}
\item 
\edit{R-Frame introduces a variable, namely epoch-wise random offset $\Delta$, based on which a dynamic training set (in frames) can be constructed in every training epoch. }

\item 
\edit{For each training epoch, bootstrapping is also applied to further improve the data diversity.}
\end{itemize}

\edit{
Given these properties (e.g., epoch-wise dynamic training-frame-generation), the proposed R-Frame can be an effective alternative to the traditional sliding window to train epoch-wise ConvNets.
Given a long signal sequence, the per-epoch training set $\pmb{X}_k$ (in the $k^{th}$ epoch) can be generated as follows:}
\begin{enumerate}
\item generate a random offset $\Delta_k\in [0, \lfloor L/2\rfloor]$, where $L$ is the window/frame length and  $\lfloor . \rfloor$ is the floor operation; 

\item remove the first $\Delta_k$ timestamps of signals from the original long sequence; 

\item perform standard sliding window approach, yielding the initial training frames;

\item bootstrapping (i.e., random sampling with replacement) on the initial training frames to form the $k^{th}$ epoch's training frames $\pmb{X}_k$.   
\end{enumerate}

Since the random offset varies over epochs, we have different epoch-wise training frames, and by applying bootstrapping(i.e., step (4)) we further enrich the data diversity for robust ConvNet training.

\edit{
In essence, R-Frame can be deemed as a special sliding window approach with additional dense-sampling strategy (i.e., based on the epoch-wise offset $\Delta$ and bootstrapping)
and it serves as the core part in the Sampling Layer of our ConvBoost framework. 
R-frame can yield various training frames at each epoch, and this characteristic makes it especially suitable for the epoch-wise ConvNet training.
}

\subsection{Mix-up Booster in Data Augmentation Layer}

For the Sampling Layer, 
we propose an R-Frame booster, which introduces epoch-wise random offset and bootstrapping strategy to generate additional training frames, in contrast to the traditional sliding window counterpart.
\edit{However, in essence it is a dense sampling approach and has its own upper limit. 
To further enrich the data diversity, we additionally use data synthesis approaches.
In our ConvBoost framework we define a
}
Data Augmentation Layer, where we use the popular mix-up strategy \cite{zhang2017mixup} as a booster to generate virtual \edit{training frames}.  

Given any two \edit{training frames} ($\pmb{x}_{i}, \pmb{y}_{i}$) and ($\pmb{x}_{j}, \pmb{y}_{j}$), the virtual \edit{frame} $\tilde{\pmb{x}}$ can be generated via a linear interpolation operation: 
\begin{equation}
    \tilde{\pmb{x}}=\lambda \pmb{x}_{i}+(1-\lambda) \pmb{x}_{j}
\end{equation}
\noindent
with two labels $\pmb{y}_{i}$ and $\pmb{y}_{j}$, where $\lambda$ is the mixing ratio sampled from Beta$(\alpha,\alpha)$ distribution\cite{zhang2017mixup} parameterized by $\alpha$, which controls the strength of interpolation.
With the generated virtual training \edit{frame}  $\tilde{\pmb{x}}$ and two labels $\pmb{y}_{i}$, $\pmb{y}_{j}$, we can calculate the corresponding joint cross-entropy loss $L^{CE}(\tilde{\pmb{x}}, \pmb{y}_{i}, \pmb{y}_{j})$ via a weighted sum operation:
\begin{equation}
L^{CE}(\tilde{\pmb{x}}, \pmb{y}_{i}, \pmb{y}_{j})=\lambda L^{CE}(\tilde{\pmb{x}}, \pmb{y}_{i})+(1-\lambda)L^{CE} (\tilde{\pmb{x}}, \pmb{y}_{j}).
\end{equation}

In this work, the mix-up booster is employed at the training batch-level, and we set the number of the virtual \edit{frames} the same as the batch size.

\subsection{Channel Dropout (C-Drop) Booster in Resilient Layer}

\begin{figure}[t]
  \centering
  \includegraphics[width=0.5\textwidth]{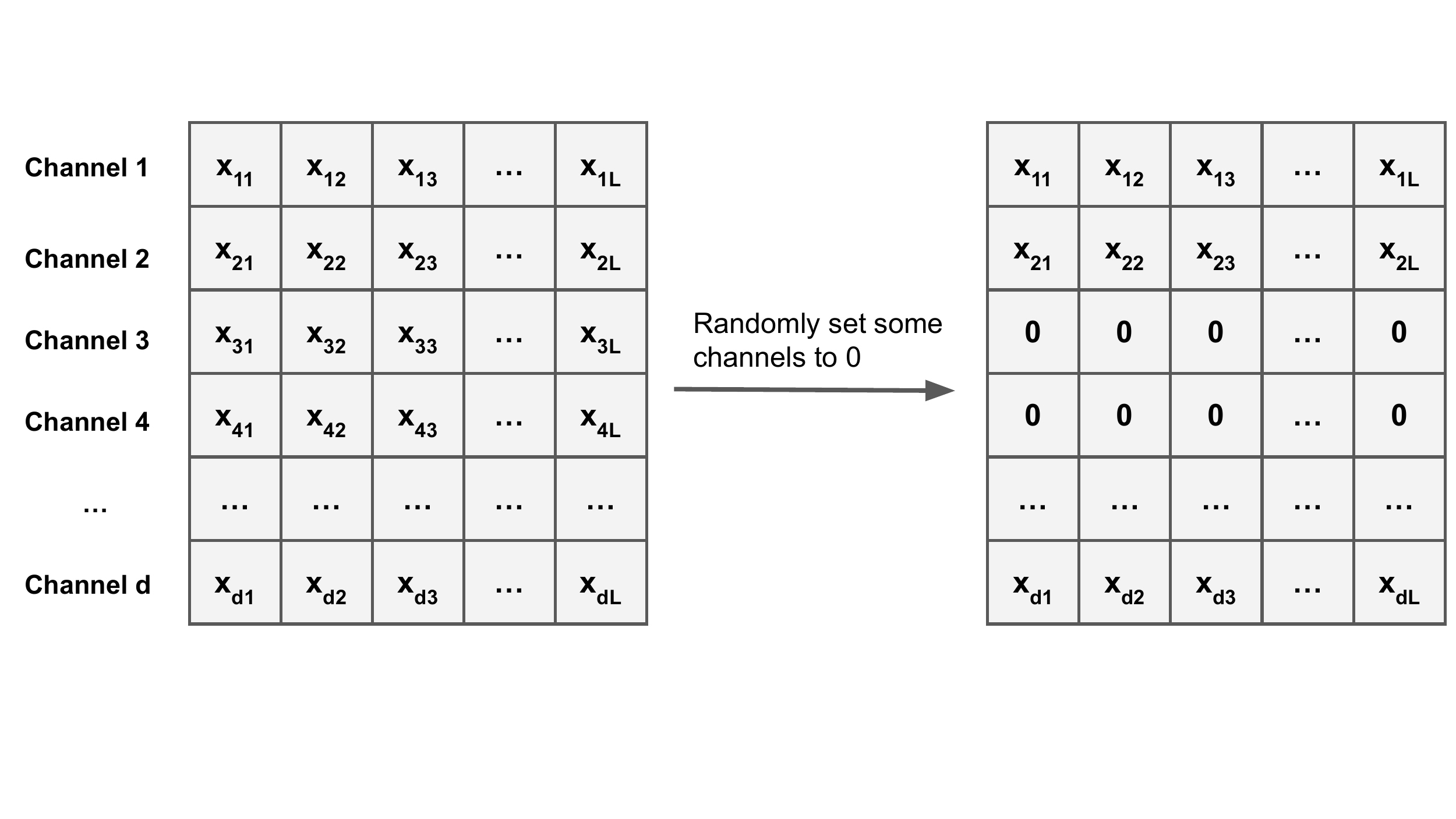}
  \caption{C-Drop booster in Resilient layer of ConvBoost framework; We randomly set some channels to 0 for diverse training data.}   
  \vspace*{-1em}
  \label{c_drop}
\end{figure}

In the Sampling and Data Augmentation Layers, 
\edit{we develop two boosters, which can create additional training frames via dense sampling and data synthesis.}
\edit{From the perspective of sensor data's characteristics,} 
it is also sensible to increase the data diversity by simulating the problematic signals, which can also make the trained model resilient to certain level of signal noises. 
\edit{Therefore, in our ConvBoost framework}, we design a Resilient Layer, and use the Channel Dropout (C-Drop) booster, which can randomly set some channels to zeros, as shown in \autoref{c_drop}.

Although dropping some channels may increase the diversity of the training data, note that the percentage of the dropped channels is a hyper-parameter and should not be too large. 
In this work, for each training batch we empirically set it as a random number ranging from $[0\%, 20\%]$, which means only a small number of channels are set to zeros. 
We expect C-Drop booster can simulate some characteristics of sensor data in real-world cases, and also increase the data diversity for robust ConvNet training.

\edit{\subsection{ConvBoost for Human Activity Recognition}
}
\edit{
\noindent
In our ConvBoost framework, we design three conceptual layers, namely Sampling Layer, Data Augmentation Layer, and Resilient Layer, based on which we develop the corresponding boosters to create more diverse training frames for robust ConvNet development.
}

\edit{
In \autoref{alg1}, we describe how to train ConvNets using proposed ConvBoost framework. 
Specifically, by applying the boosters in the proposed layers, we can generate per-epoch training frames, based on which epoch-wise HAR classifiers can be trained. 
Without loss of generality, our framework outputs all the epoch-wise classifiers for HAR tasks, as shown in \autoref{alg1}.
}
It is a flexible framework and can be used directly \edit{by selecting the best epoch-wise classifier based on validation--\emph{Single-Best}--or via \emph{Ensemble} \cite{guan2017}.
Both schemes are studied in the experimental section.
}

\edit{For the Ensemble scheme, similar to \cite{guan2017}, based on validation data, we can choose $M$ best base learners $\{ \pmb{W}^m \}_{m=1}^M$ for aggregation.
}
For a query data $\pmb{x}$, the classification probability distribution of the $m^{th}$ model can be written as $p(\pmb{y}|\pmb{x}; \pmb{W}^m)$, and via a simple score-level fusion we can further get the final aggregated score $p(\pmb{y}|\pmb{x})$:
\begin{equation}
p(\pmb{y}|\pmb{x}) = \frac{1}{M} \sum_{m=1}^{M} p( \pmb{y} | \pmb{x}; \pmb{W}^m),
\end{equation}
and predicted activity $\hat{y}$  will be assigned to the one with the largest probability over the $C$ classes, i.e., $\hat{y} = \arg \max_{\{1,...,C\}}p(\pmb{y}|\pmb{x})$.\\

\RestyleAlgo{ruled}
\SetKwComment{Comment}{/* }{ */}
\SetKwInput{KwInput}{Input}  
\SetKwInput{KwOutput}{Output} 

\begin{algorithm}[htbp]
\caption{\edit{ConvBoost for HAR Model Training}} 

\KwInput{Original long sequence}
\KwOutput{Epoch-wise ConvNet models $\{ \pmb{W}^k \}_{k=1}^K$, where $K$ is the total training epochs}
Model Initialization  \\
\For {$k = 1$ to $K$}{
  \emph{Sampling Layer}: Applying R-frame Booster (Sec. 3.1) \\
  Shuffling and Constructing $B$ batches\\
  \For {$b = 1$ to B}{
  \emph{Resilient Layer}: Applying C-Drop Booster (Sec. 3.3) \\
  \emph{Data Augmentation Layer}: Applying Mix-up Booster (Sec. 3.2) \\
  Model Training (via Backpropagation) 
  }
  Output epoch-wise model $\pmb{W}_{}^{k}$ 
}
\label{alg1}
\end{algorithm}

%% file: sections/experiment.tex
\subsection{Datasets} 
\begin{figure}[t]
  \centering
  \includegraphics[width=\textwidth]{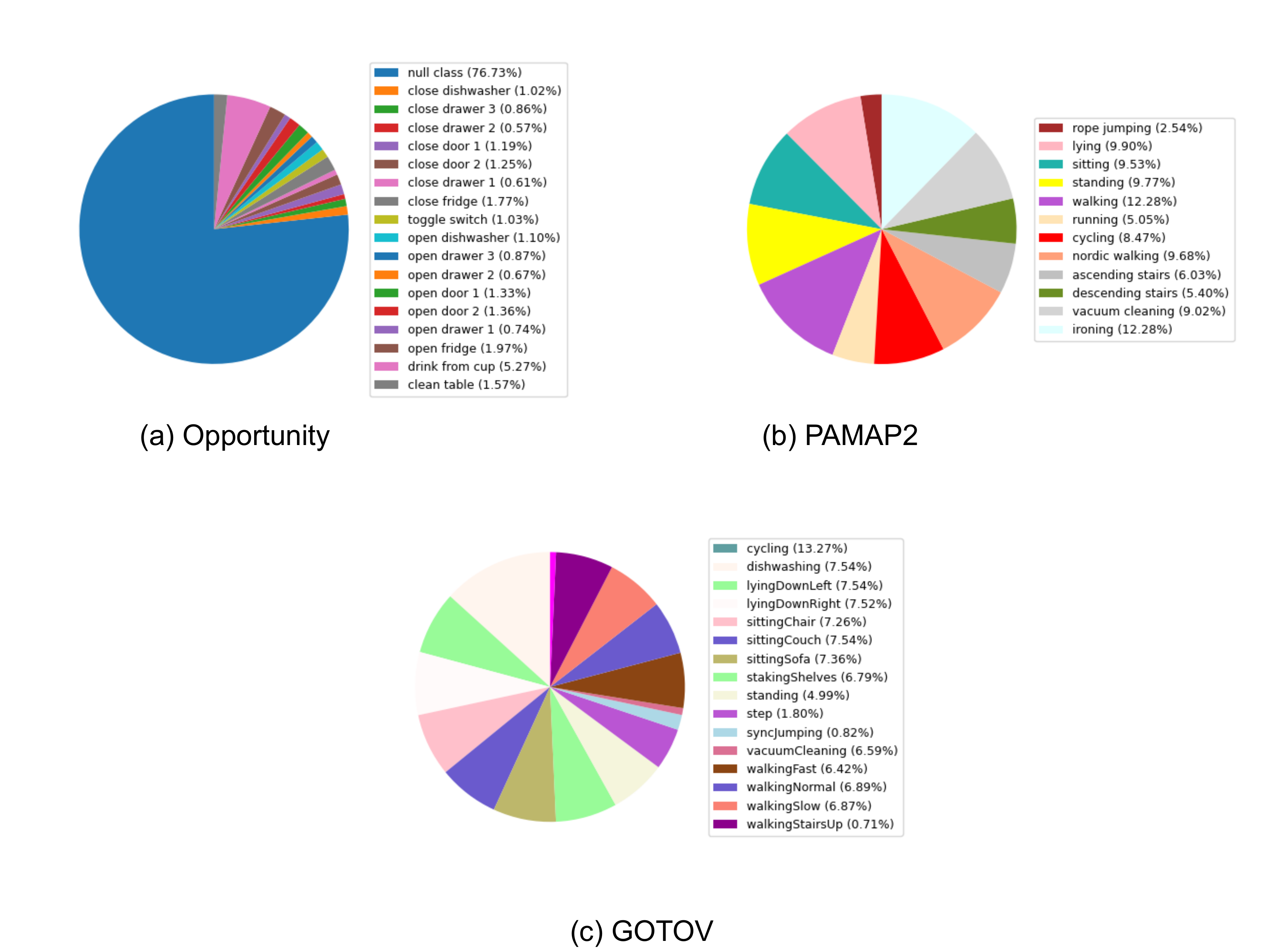}
  \caption{Activity distributions of Opportunity, PAMAP2, and GOTOV datasets.}
  \label{dataset_pie_chart}
\end{figure}

To evaluate the effectiveness of our method, we used three public datasets: 
\emph{i)} Opportunity (OPP) \cite{chavarriaga2013opportunity};
\emph{ii)} Physical Activity Monitoring (PAMAP2) dataset \cite{reiss2012introducing}; and
\emph{iii)} Growing Old Together Validation (GOTOV) \cite{paraschiakos2020activity} dataset.
\edit{The activities to be recognized}
for each dataset are shown in \autoref{dataset_pie_chart}.

\subsubsection{Opportunity (OPP) \cite{chavarriaga2013opportunity}} 
The OPP dataset is one of the most challenging wearable-based HAR dataset, which exhibits imbalanced class distributions (as shown in \autoref{dataset_pie_chart}).
It includes 18 daily kitchen activities (collected from five runs of four subjects) such as opening the door or closing the drawer, at a sampling rate of 30 Hz. 
Following \cite{hammerla2016deep,guan2017}, 
we employed the hold-out evaluation protocol.
That is, the second run from subject 1 was used for validation, runs 4 and 5 from subjects 2 and 3 were used as test while the rest data were used for training. Following \cite{hammerla2016deep,guan2017}, 79-dimensional IMU recordings were used for our experiments.

\subsubsection{Physical Activity Monitoring Dataset (PAMAP2) \cite{reiss2012introducing}}
The PAMAP2 dataset is one of the most widely used wearable-based HAR dataset, which includes 12 daily activities (collected from nine subjects) such as 
running, walking, lying, sitting, etc. 
The dataset includes IMU recordings from hand, chest and ankle with accelerometer, gyroscope, magnetometer, temperature and heart rate information, with a total of 52 dimensions.
We employed the hold-out evaluation protocol from \cite{hammerla2016deep,guan2017}, that is, the runs 1 and 2 from subject five were used for validation, runs 1 and 2 from the sixth subject were used as test, while the remaining data was used for training. 

\subsubsection{Growing Old Together Validation (GOTOV) \cite{paraschiakos2020activity}}
GOTOV is one of the most recent HAR dataset, with activities collected from \edit{older-age subjects}. 
It has 16 daily activities collected from thirty-five  participants.
The subjects were instructed to wear accelerometer sensors at three locations: ankle, chest and wrist with a 9-dimensional recording, at the sampling rate of 83Hz
\cite{paraschiakos2020activity}. 
In our study, we removed six participants who did not wear the sensors completely and used the rest twenty-nine subjects for our experiments. 
Hold-out evaluation was employed and we used data from subjects 9, 13, 20, 30, 31 for validation; subjects 6, 17, 28, 33, 35 for testing and the rest nineteen subjects for training.

\subsection{Implementation Details}

\subsubsection{Networks and Configurations}
To boost ConvNets, we employed some most popular backbones: CNN\cite{yang2015deep}, convolutional LSTM (ConvLSTM)\cite{morales2016deep} and attention model\cite{Murahari2018}. 
For CNN, three convolutional layers (with 256 feature maps per layer)
were used, with max-pooling layers in-between, 
followed by two fully-connected layers (with hidden unit number 128 per layer) connecting to the output nodes.
For ConvLSTM, three convolutional layers (with 256 feature maps per layer) were used, with max-pooling layers in-between,  followed by two LSTM layers (with hidden unit number 128 per layer) connecting to the output nodes. 
For Attention Model, four convolutional layers (with 64 feature maps per layer) were used, followed by two LSTM layers (with 128 hidden units per layer)\cite{ordonez2016deep} and an attention layer \cite{Murahari2018}.
\edit{For all hidden units in the three backbone networks, ReLU was used as the activation function. 
}

To train the models, cross entropy loss with Adam optimizer was used.
Each model was trained for 100 epochs, and 100 base classifiers were generated via \autoref{alg1}. 
During training, group normalization \cite{Wu2020} was used, and we set the mini-batch size to 256; dropout was performed before the output layer with $50\%$;  
The learning rate was set to ${10^{-3}}$, and we fixed $\alpha$=0.8 for mix-up method.
These hyper-parameters were used for all ConvNets across all the datasets.

For OPP, following \cite{hammerla2016deep}, the length of the sliding window was set to 1 second, with 50\% overlap. 
We also used 1 second sliding window with 50\% overlap on GOTOV dataset.
For PAMAP2 dataset, following \cite{hammerla2016deep}, we used sliding window in the length of 5.12 seconds with 78\% overlap.
Following \cite{guan2017}, we normalized each dataset before training/evaluation, i.e., making each channel zero mean and unit variance.

\subsubsection{Evaluation Metric}
For all experiments, mean F1-score was used to measure the performance, which is defined as:
\begin{equation}
\bar{F}_{1}=\frac{1}{C} \sum_{c=1}^{C} \frac{2 \mathbf{T} \mathbf{P}_{c}}{2 \mathbf{T} \mathbf{P}_{c}+\mathbf{F} \mathbf{P}_{c}+\mathbf{F} \mathbf{N}_{c}},
\end{equation}
where $C$ stands for the number of classes (activities); For the $c^{th}$ class, $\mathbf{T}\mathbf{P}_{c}$, $\mathbf{F}\mathbf{P}_{c}$, $\mathbf{F}\mathbf{N}_{c}$ denote the number of true positive, false positive, and false negative predictions, respectively.
For the statistical significance testing, we used two-tailed independent t-tests with p-values reported, where
p $\leq$ 0.05, p $\leq$ 0.01 and p $\leq$ 0.001 correspond to *, ** and ***, respectively (in line with the literature e.g., \cite{morales2016deep},\cite{guan2017}).

\subsection{Model Comparison}
Based on the OPP, PAMAP2 and GOTOV datasets, we evaluated our \edit{ConvBoost} framework using state-of-the-art ConvNet backbones from the HAR community:
\emph{i)} CNN \cite{yang2015deep};
\emph{ii)} ConvLSTM \cite{morales2016deep}; and 
\emph{iii)} Attention Model \cite{Murahari2018}.

\edit{
Our ConvBoost framework is motivated by previous work \cite{guan2017}, where per-epoch generated classifiers (i.e., sample-wise stateful LSTMs) were fused for improved HAR. 
For better fusion effect, hyper-parameters such as window length, batch size, initial sampling point were modelled as random factors to enhance the ensemble diversity.
Although the frame-wise ConvNets are different from sample-wise LSTM, based on the idea of epoch-wise bagging and diversity injection, we developed the ConvNet variant of \cite{guan2017} for comparison, with the implementation details as follows: 
}

\begin{table}[t]
  \centering
    \caption{\edit{Results comparison for our proposed ConvBoost and the ConvNet variant of \cite{guan2017} with different backbones on three public datasets;  
    The $\bar{F}_{1}$ results (in $\%$, with mean and standard deviation of 20 repetitions) are reported for both Single-Best/Ensemble modes}.}
    \label{result_all}
    \begin{tabular}{c|l|c|c|c}
    \toprule
    \multicolumn{2}{c|}{Methods} & OPP & PAMAP2 & GOTOV \\
    \midrule
    \multirow{5}{*}{CNN\cite{yang2015deep}} & Original & 61.2 & 81.0 & 74.6 \\
     & \edit{ConvNet variant of \cite{guan2017}}(Single-Best)  & 61.51 $\pm$ 1.51 & 82.03 $\pm$ 3.99 & 74.79 $\pm$ 1.83 \\
    & \edit{ConvNet variant of \cite{guan2017}}(Ensemble) & 65.61 $\pm$ 1.23 & 85.25 $\pm$ 3.41 & 76.67 $\pm$ 1.21 \\
    &  \edit{Proposed ConvBoost }(Single-Best) & 69.10 $\pm$ 1.57 & 89.92 $\pm$ 1.79 & 79.51 $\pm$ 1.33 \\
   & \edit{Proposed ConvBoost }(Ensemble) & \textbf{70.53 $\pm$ 0.81} & \textbf{90.05 $\pm$ 0.56} & \textbf{80.99 $\pm$ 0.88} \\
    \midrule
    \multicolumn{1}{c|}{\multirow{5}{*}{ConvLSTM\cite{morales2016deep}}} & Original & 62.2 & 77.7 & 72.0  \\
    & \edit{ConvNet variant of \cite{guan2017}}(Single-Best) & 62.24 $\pm$ 1.67 & 80.03 $\pm$ 3.07 & 72.36 $\pm$ 2.67 \\
    & \edit{ConvNet variant of \cite{guan2017}}(Ensemble)& 65.50 $\pm$ 1.11 & 83.47 $\pm$ 3.49 & 74.33 $\pm$ 1.56 \\
    & \edit{Proposed ConvBoost }(Single-Best) & 68.22 $\pm$ 1.99 & 89.37 $\pm$ 1.90 & 77.44 $\pm$ 1.51 \\
    & \edit{Proposed ConvBoost }(Ensemble) & \textbf{71.24 $\pm$ 0.96} & \textbf{90.26 $\pm$ 0.40} & \textbf{78.95 $\pm$ 1.01} \\
    \midrule
    \multicolumn{1}{c|}{\multirow{5}{*}{Att. Model\cite{Murahari2018}}} & Original &  64.1 & 88.1  & 72.4 \\
     & \edit{ConvNet variant of \cite{guan2017}}(Single-Best) & 64.05 $\pm$ 1.06 & 88.58 $\pm$ 3.41 & 72.24 $\pm$ 1.89 \\
    & \edit{ConvNet variant of \cite{guan2017}}(Ensemble) & 67.58 $\pm$ 1.23 & 89.88 $\pm$ 2.95 & 74.68 $\pm$ 1.51 \\
   & \edit{Proposed ConvBoost }(Single-Best) & 71.05 $\pm$ 1.18 & 89.57 $\pm$ 2.21 & 80.08 $\pm$ 0.96 \\
    & \edit{Proposed ConvBoost }(Ensemble) & \textbf{72.81 $\pm$ 0.76} & \textbf{89.90 $\pm$ 0.57} & \textbf{81.66 $\pm$ 0.72} \\
    \midrule
    \end{tabular}%
\end{table}%

\begin{itemize}
\item 
\edit{\emph{ConvNet variant of \cite{guan2017} (Ensemble):} The epoch-wise bagging strategy was applied to the (frame-wise) ConvNet backbone. 
For each training epoch, we applied bootstrapping for diversity injection, i.e., we randomly sample the frames with replacement, which will result in about 36.8\% \cite{chernick2014introduction} of the frames not being used (per epoch), yielding diverse epoch-wise base classifiers. 
In the Ensemble mode, following \cite{guan2017}, the 20 best epoch-wise classifiers were selected (via validation set) for score-level fusion.
}

\item 
\edit{\emph{ConvNet variant of \cite{guan2017} (Single-Best):} This is a special case of the Ensemble mode, where only the best base classifier was selected (via validation set) for HAR.
}
\end{itemize}

\edit{
In both Ensemble and Single-Best modes, we compared the ConvNet variants of \cite{guan2017} with our ConvBoost framework:  
}

\begin{itemize}
\item \emph{Proposed \edit{ConvBoost (Ensemble)}:} 
\edit{The proposed 3-layer framework with the following three} boosters, R-Frame, Mix-up and C-Drop;
\edit{Likewise, in the Ensemble mode, the 20 best epoch-wise classifiers were selected (via validation set) for score-level fusion.
It is worth noting that ConvNet variant of \cite{guan2017} is a special case of our ConvBoost when without boosters, i.e., without additional generated training frames. 
}

\item 
\emph{Proposed \edit{ConvBoost (Single-Best)}:} Only the best base classifier within the \edit{ConvBoost Ensemble was used (via validation set) for HAR.}

\end{itemize}

\begin{figure}[t]
  \centering
  \includegraphics[width=\textwidth]{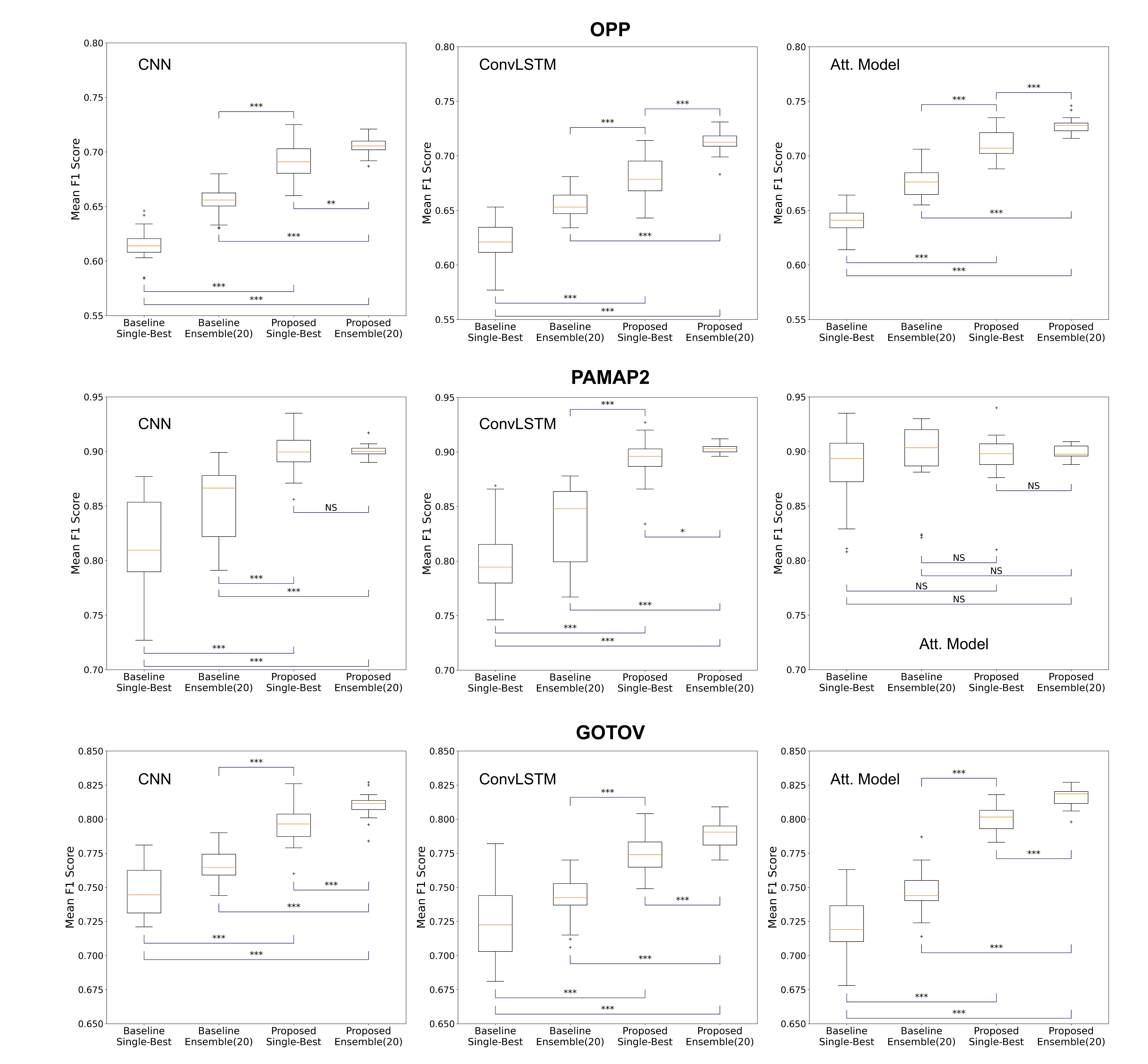}
  
   \caption{ \edit{Results of experimental evaluations for the four methods assessed: Single-Best/Ensemble modes of the proposed ConvBoost as well as the ConvNet variant of \cite{guan2017}(referred to Baseline here) with three different backbone ConvNets (CNN, ConvLSTM, Att.\ Model) on three public datasets (OPP, PAMAP2, and GOTOV); NS denotes not significant. }
   }
  
  \label{box_plot}
\end{figure}

Based on the aforementioned ConvNet backbones (CNN \cite{yang2015deep}, ConvLSTM \cite{morales2016deep}, and Attention Model \cite{Murahari2018}), 
we report the $\bar{F}_1$ results of:
\emph{i)} the Proposed ConvBoost;
\emph{ii)} ConvNet variants of \cite{guan2017}; as well as 
\emph{iii)} the original ConvNet baselines in \autoref{result_all}.
For the Proposed ConvBoost  and the ConvNet variants of \cite{guan2017}, both the results of Ensemble and Single-Best modes are reported.
Since there are random factors in the training process,
we ran each model for 20 repetitions and report the corresponding mean and standard deviation.

From \autoref{result_all}, we can see--compared to the original backbone ConvNets--generally it is \edit{ beneficial to apply the epoch-wise bagging strategy \cite{guan2017} to ConvNets. 
Based on the developed ConvNet variant of \cite{guan2017}, we can see although Single-Best only} has comparable performance, substantial performance gain can be achieved \edit{in the Ensemble mode. 
}

\edit{
However, the potential of the epoch-wise bagging scheme may not be fully exploited when with inadequate, i.e., too little, training data. 
Our ConvBoost framework can provide an simple yet effective solution, based on which the additional training frames can be generated via three conceptual layers, i.e., Sampling Layer, Data Augmentation Layer, and Resilient Layer. 
From \autoref{result_all}, we can see with the generated additional training data, 
significant performance improvements can be achieved, when compared to the Single-Best/Ensemble counterparts of ConvNet varants of \cite{guan2017},   
}
on all the three datasets, irrespective of backbone ConvNets. 
\edit{It is worth noting that based on our ConvBoost, single classifier (i.e., Single-Best) can outperform classifier fusion method (i.e., Ensemble of ConvNet variant of \cite{guan2017}), especially in the challenging datasets (e.g., OPP, and GOTOV), indicating effectiveness of our ConvBoost framework, which can fully take advantage of the epoch-wise bagging scheme via generating additional training frames via its three different conceptual layers.    
}

We also conducted the statistical significance testing \edit{on the four methods (i.e., Single-Best and Ensemble modes of our ConvBoost and ConvNet variant of \cite{guan2017})}, and the corresponding box-plots are provided in \autoref{box_plot}. 
\edit{It becomes clear that the results are in line with the previous observations in \autoref{result_all}.}
From \autoref{box_plot} we notice for the PAMAP2 dataset, with Attention Model backbone, all the results of the four methods are not significantly different, 
with $\bar{F_1}$ (in $\%$) ranging from $88.58\pm 3.41$ (Single-Best \edit{of ConvNet variant of \cite{guan2017}}) to $89.90\pm0.57$ (\edit{Proposed ConvBoost Ensemble}).
One possible explanation can be the high-performance of the backbone (88.1$\%$ in $\bar{F_1}$ of Attention Model on PAMAP2 dataset), which may hit the upper limit of the performance on this dataset\footnote{More results can be found in Appendix including a) leave-one-subject-out setting on PAMAP2 dataset ( \autoref{loso_pamap2}); b) Confusion matrices ( \autoref{confusion_matrix_att_model})}. 
On the more challenging datasets (i.e., OPP and GOTOV datasets), the performance gains are significant for our approach \edit{irrespective of Single-Best or Ensemble mode}. 
For example, with Attention Model as backbone, \edit{our ConvBoost (Ensemble)} leads to 
about $7\%$, and $5\%$ performance gains over \edit{the ConvNet variant of \cite{guan2017} (Ensemble)} on GOTOV dataset and OPP dataset respectively, indicating the effectiveness of the proposed \edit{ConvBoost framework} on challenging HAR scenarios.   

\edit{
Generally, for the epoch-wise training schemes, i.e., our ConvBoost or the ConvNet variant of \cite{guan2017}, we can observe the Ensemble mode tends to perform better than the Single-Best counterpart, indicating the generalization capability of classifier fusion. 
}
\edit{ 
One the other hand, our 3-layer ConvBoost framework can generate additional informative training frames, based on which even single classifiers can  also yield very competitive performance. 
As shown in \autoref{box_plot}, the Single-Best (ConvBoost) can benefit significantly from the additional training data, and it can substantially outperform Ensemble when without such generated data (i.e., ConvNet variant of \cite{guan2017}). 
This observation suggests the significant contribution of the training-frame-generation mechanism of our ConvBoost framework via 3 different layers/boosters, suggesting it is an important direction for robust HAR model development.
}

\edit{Since the Ensemble mode of our ConvBoost framework can yield the best performance, unless stated otherwise, we use it as the default model for the rest of this paper.
}

\subsection{Ablation Study}
\edit{Based on the ConvBoost Ensemble, we also conducted a number of ablation studies to:
\emph{i)} better understand the performance gain contributed by each booster (in the corresponding layer); and
\emph{ii)} better understand how booster can change the properties of the ensemble in terms of diversity and expected performance of base classifiers}.
Note all the experiments in this subsection were conducted on the OPP dataset.

\begin{figure}[htbp]
\centering
\includegraphics[width=0.32\textwidth]{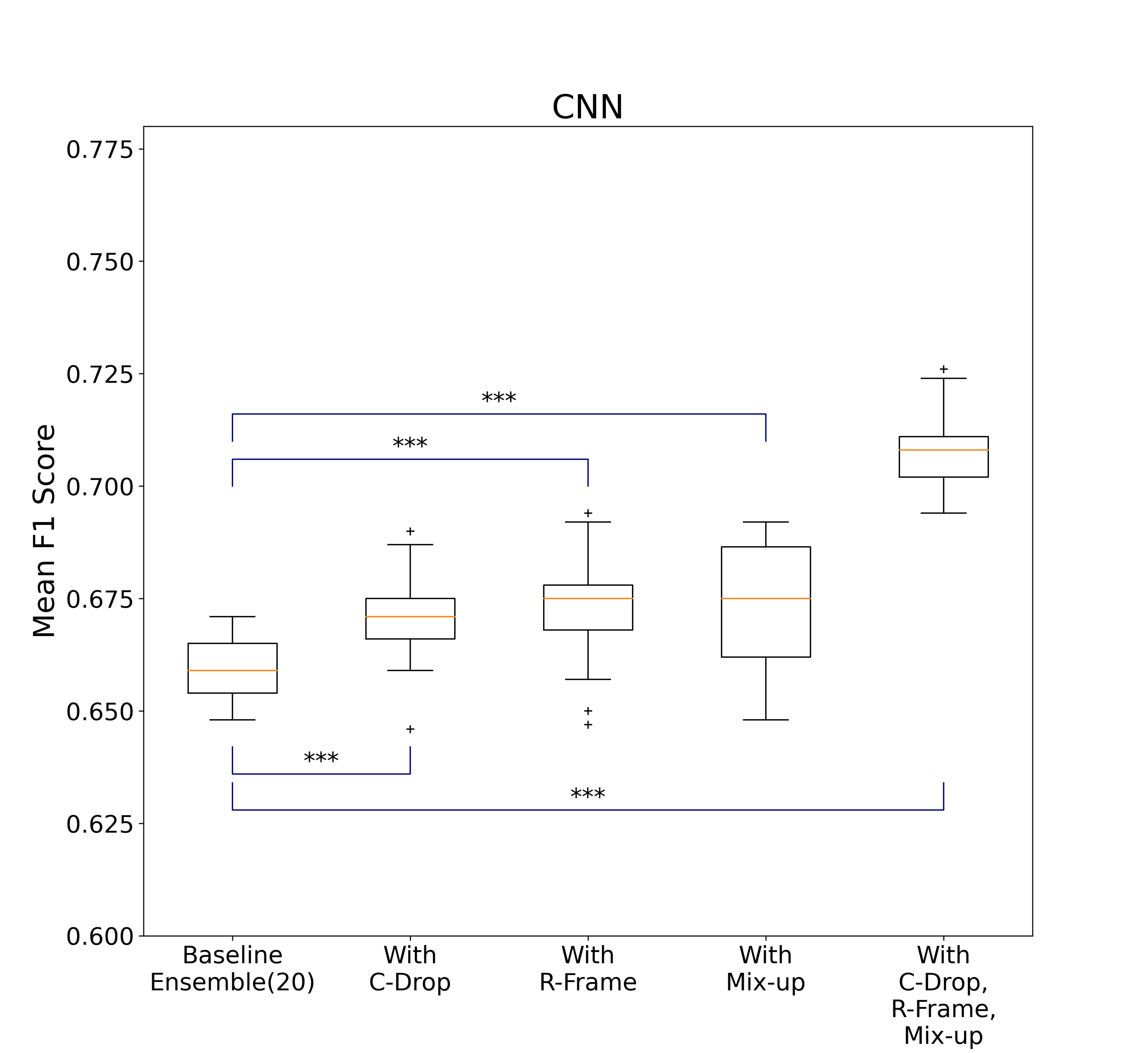}
\includegraphics[width=0.32\textwidth]{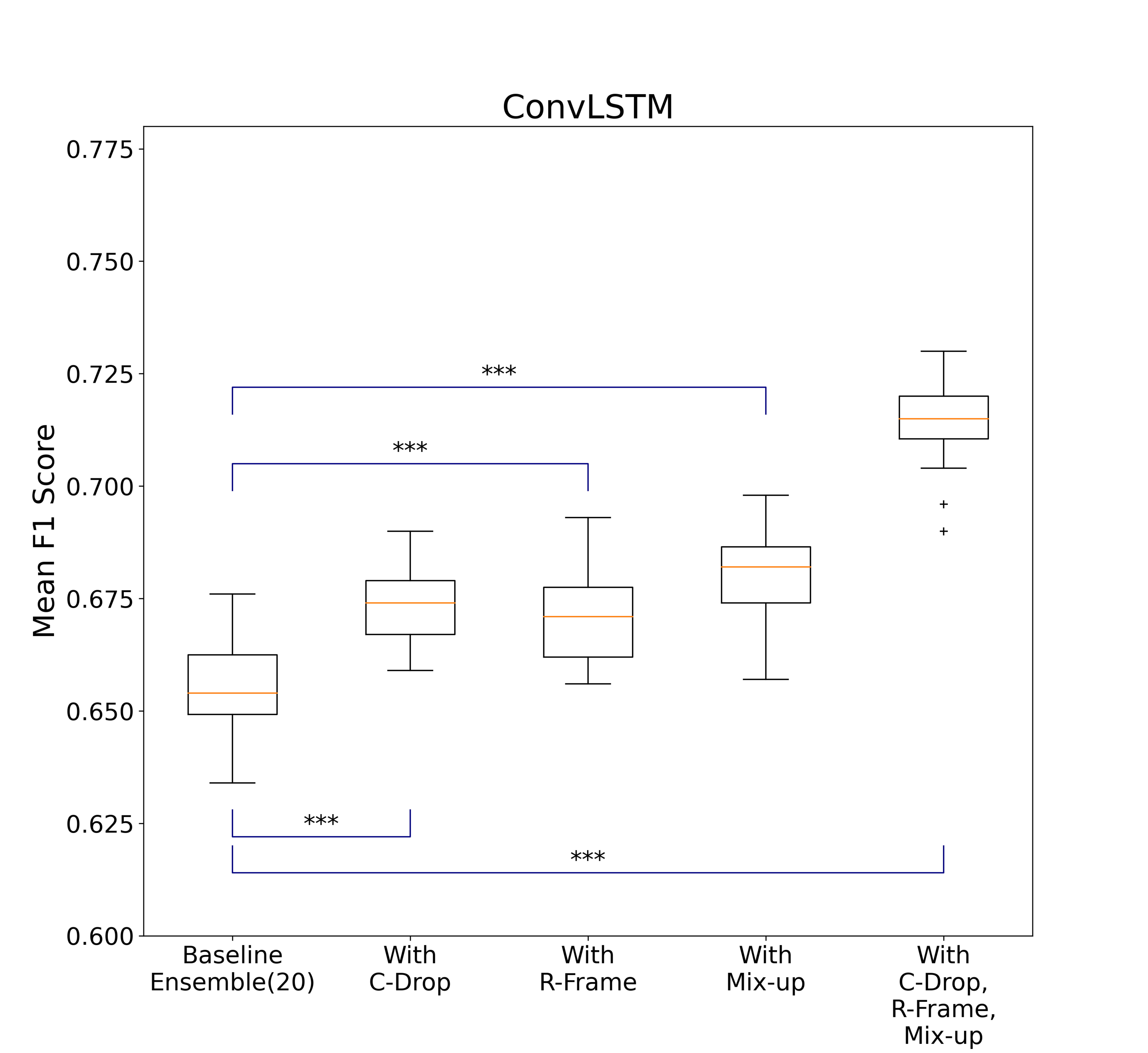}
\includegraphics[width=0.32\textwidth]{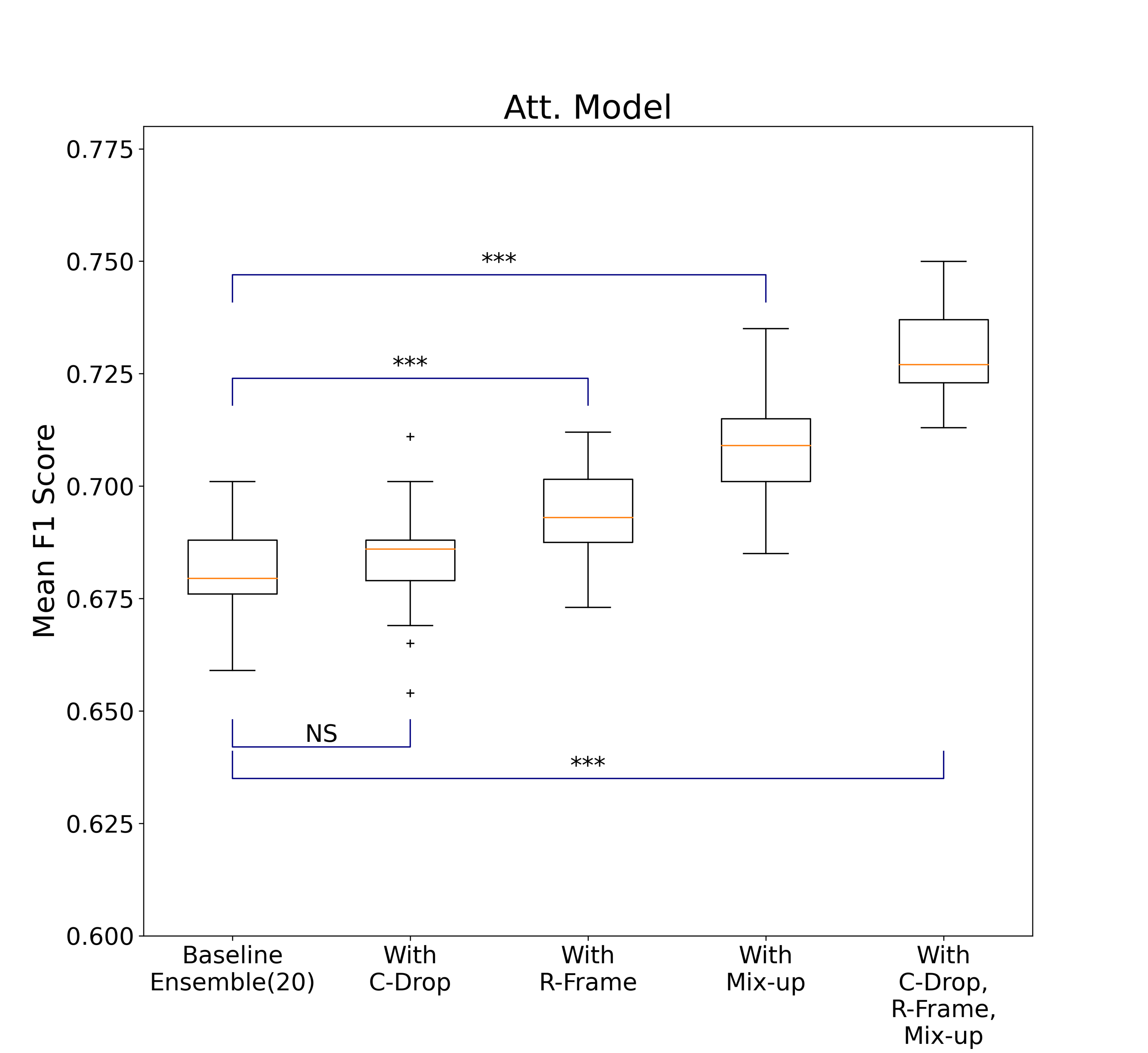}
  \caption{Box-plots of ablation studies on booster(s) on OPP dataset; Baseline Ensemble (i.e., ConvNet variant of \cite{guan2017}) includes no boosters, and our proposed ConvBoost Ensemble includes all 3 boosters, i.e., R-Frame, Mix-up, C-Drop
  }
\label{boxplot_ablation}
\end{figure}

\subsubsection{On the Effectiveness of Boosters}
\edit{In our 3-layer ConvBoost framework, the three boosters (R-Frame, Mix-up, and C-Drop) are the core components, aiming at generating various training frames at different stages.
To better understand the contribution of each booster, we designed some ablation studies and report the results in form of box-plots results in \autoref{boxplot_ablation}.
}
We can observe that the proposed \edit{ConvBoost Ensemble} with all the 3 boosters (i.e., with R-Frame, Mix-up, and C-Drop) yields the best results, much higher than \edit{Baseline Ensemble (i.e., ConvNet variant of \cite{guan2017}, with no booster)} or the ones with single booster, indicating the complementary nature of these boosters. 
\edit{
The three boosters can generate additional training data from very different perspectives, e.g., via dense-sampling in Sampling Layer, via interpolation/synthesis in Data Augmentation Layer, and via sensor data simulation in Resilient Layer, respectively. 
These different types of booster-generated data tend to be less correlated, yielding very promising combining results when used as a whole.
}

\edit{It can also be observed that the application of single booster can generally} improve the performance significantly irrespective of ConvNets. 
The only exception is the Attention Model backbone with C-Drop booster, which has comparable performance with the Baseline Ensemble (i.e., no booster). 
Nevertheless, it is still beneficial to use other two boosters (i.e., R-Frame and Mix-up) \edit{with improved} results.

\begin{figure}[htbp]
\centering
\includegraphics[width=0.32\textwidth]{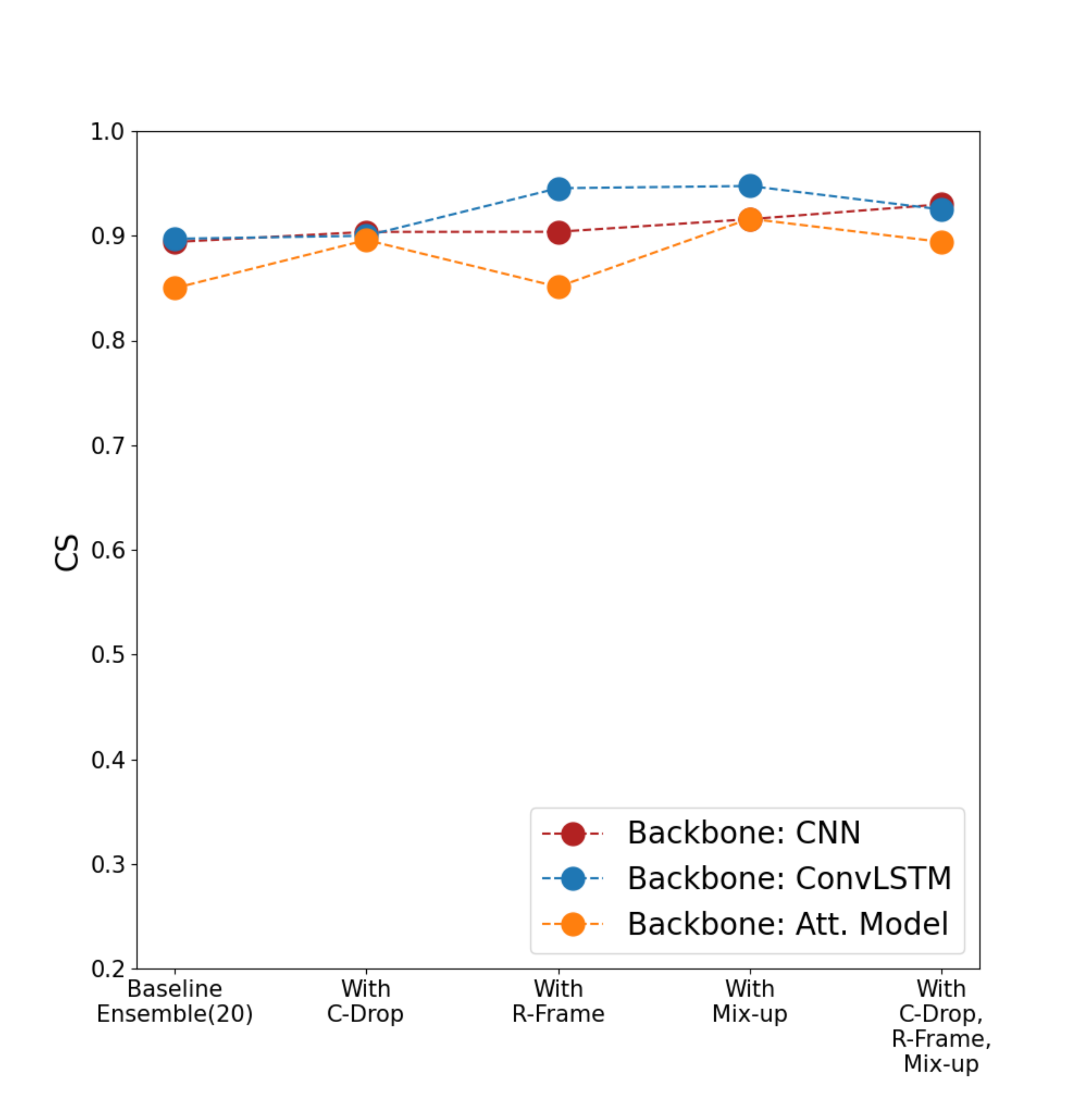}
\includegraphics[width=0.32\textwidth]{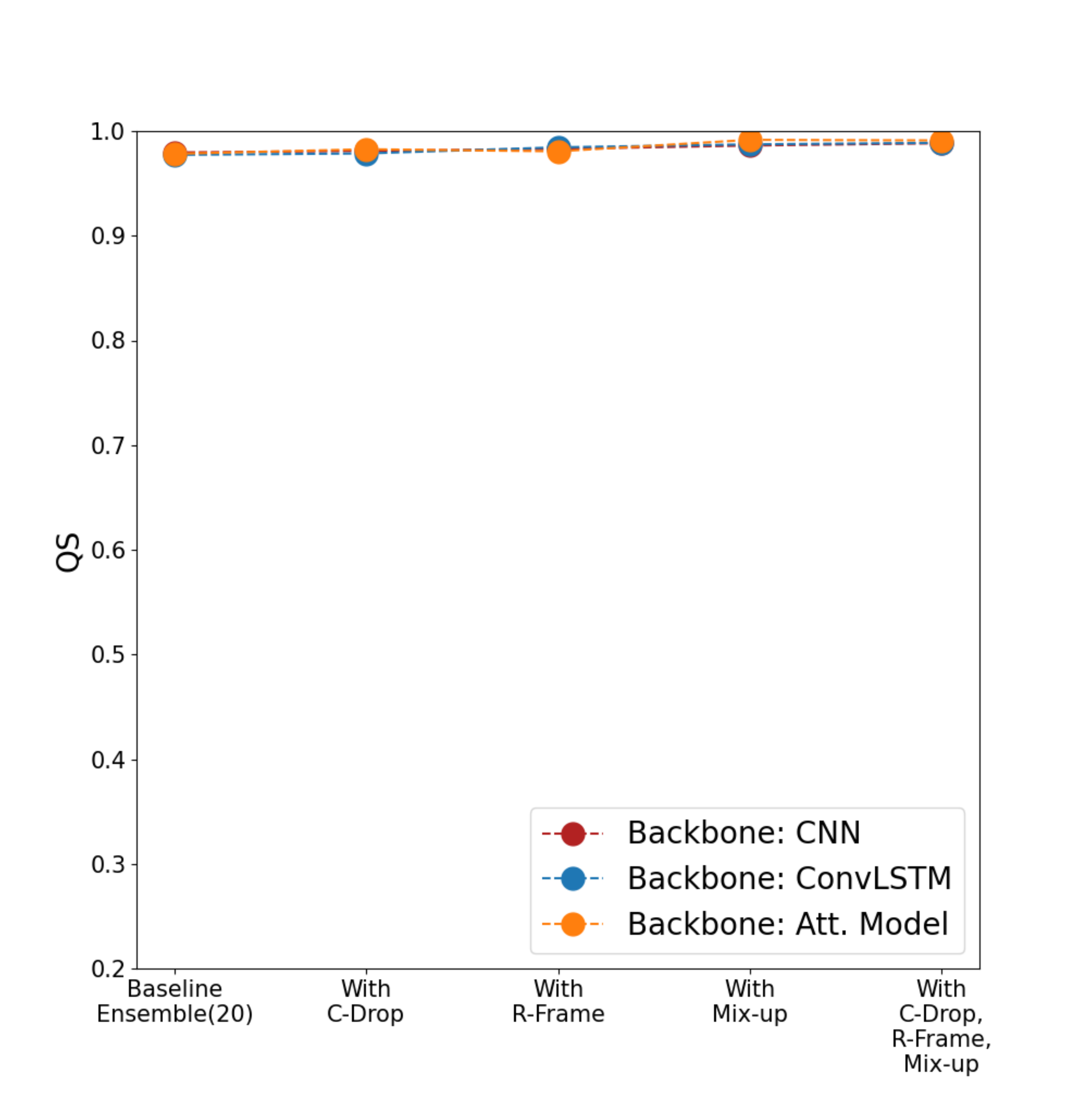}
\includegraphics[width=0.32\textwidth]{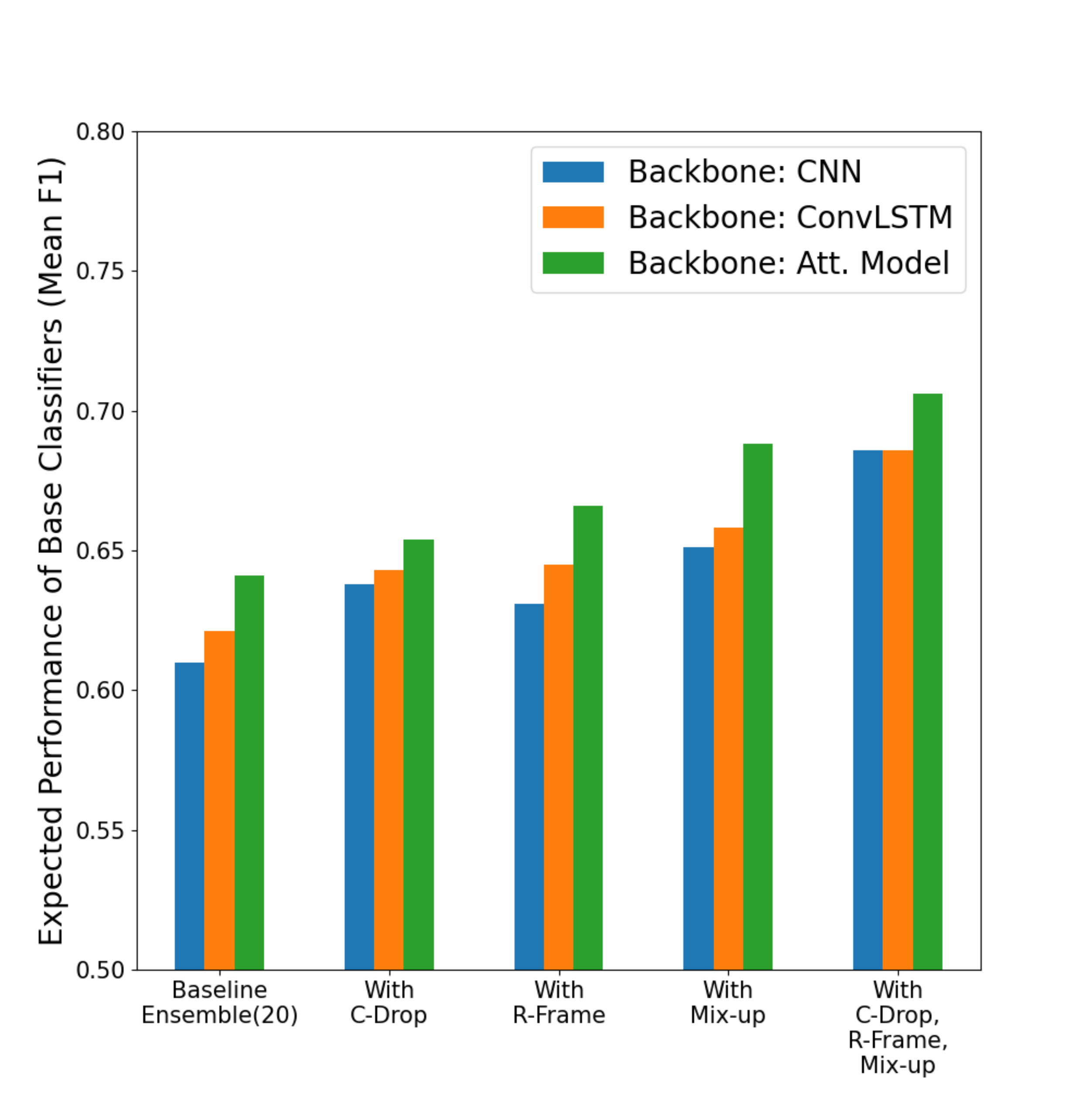}
  \caption{Ensemble diversity distributions in different ensemble settings using two metrics: CS(left) and QS(middle); Right: Expected $\bar{F}_1$ of base classifiers in different ensemble settings; Note these results are based on OPP dataset, and higher values of CS or QS indicate lower diversity.
  Baseline Ensemble (i.e., ConvNet variant of \cite{guan2017}) includes no boosters, and our proposed ConvBoost Ensemble includes all 3 boosters, i.e., R-Frame, Mix-up, C-Drop
  }
  \label{cs_qs_result}
\end{figure}

\subsubsection{Ensemble Analysis}
For ensembles, generally higher level of diversity (among the base classifiers) tends to yield better fusing results \cite{zhou2019ensemble, liu2019deep, jackowski2018new, kuncheva2003measures}.   
To measure the diversity, in this work we use two metrics, namely, the Q-Statistic (QS)\cite{udny1900association}, and cosine-similarity of the model parameters (i.e., CS).
QS exploits the predicted labels (from the base classifier pairs) to measure diversity, while CS directly measures the similarity of the pair-wise model parameters without any prediction process. More details of QS can be found in the Appendix.
For both metrics, higher values (with a maximum value of 1) indicate lower diversity. 

The performance of base classifiers is also key to the ensemble, and previous works suggested ensemble's performance can be boosted by strengthening base classifiers \cite{guan2013combining, guan2014human}. 
Here we measure both diversity (in CS/QS) and base classifiers' expected performance (in $\bar{F}_1$) and report the results in \autoref{cs_qs_result}. 
We can see the CS/QS values are constantly high, indicating lower diversity irrespective of boosters or backbones.
One possible explanation is that the models were generated in an epoch-wise manner, resulting in higher level of similarity (with lower diversity). 
One alternative to increase the ensemble diversity is to employ different network structures, e.g., fusing CNN, LSTM, or ConvLSTM, etc.\ yet \edit{it is less practical since}
the training cost can be much higher than epoch-wise training strategies.  

From \autoref{cs_qs_result} we can also see for all the three ConvNet backbones, the base classifiers can be strengthened by the three boosters, and the expected performance (in $\bar{F}_1$) of base classifiers is in line with the corresponding ensemble performance (as shown in \autoref{boxplot_ablation}).

These observations suggest the performance gains are mainly from the strengthened base classifiers, instead of ensemble diversity.
\edit{This finding can explain the superior performance of our ConvBoost's Single-Best mode, whose performance are only slightly lower the the corresponding Ensemble mode (as shown in \autoref{result_all}), and it highlights the importance of training data generation under the proposed framework.  
}

\subsection{\edit{Extensions of the ConvBoost Framework}}

\edit{The proposed ConvBoost is an extensible framework, and in this subsection we demonstrate two possible extensions by:
\emph{i)} adding extra boosters; and 
\emph{ii)} compressing ConvBoost Ensemble. 
}

\subsubsection{Extra Booster(s)}
\edit{In our 3-layer ConvBoost framework, three boosters are used to generate training frames from different perspectives.
Experimental results in the ablation studies suggest their complementary nature and effectiveness of combining them.
It is also interesting to explore extra boosters for improved performance, and based on our 3-layer ConvBoost structure, we designed experiments with simple additional boosters:
}

\begin{enumerate}
\item 
\edit{\emph{Experiment 1:} We additionally add R-Frame$^{*}$, i.e., R-Frame followed by Mix-up in the Data Augmentation Layer.
}

\item 
\edit{\emph{Experiment 2:} we additionally add scaling operation \cite{um2017data} in the Resilient Layer.
}

\end{enumerate}

\edit{
Based on the OPP dataset, in \autoref{extension_table} we compare their performance with Baseline Ensemble (i.e., ConvNet variant of \cite{guan2017}, no booster) and the proposed ConvBoost Ensemble (with 3 boosters, or different combinations of 2 boosters). 
}
We can see with these simple extra boosters, the performance can be further improved to some extent, suggesting the extensibility of the ConvBoost framework. 
To further boost the performance,
one may design more complementary boosters or layers that are less correlated to the existing ones under this ConvBoost framework. 
For example, most recently it was found that context information can also be combined into frame-wise ConvNets \cite{roggen20} as additional information, and in the future we will incorporate this context-aware idea into our ConvBoost framework for booster/layer design.

\edit{
From \autoref{extension_table},
it is also worth noting that within our ConvBoost framework, the ConvNet variant of \cite{guan2017} (referred to Baseline Ensemble in the table) can be deemed as a special case when with no boosters. 
Without additional generated training data, it has lowest performance than other ones with more boosters. 
}

\begin{table}[t]
    \centering
    \caption{Performance ($\bar{F}_1$) Distribution of \edit{the ConvBoost} framework with extra boosters on OPP dataset; R-Frame$^{*}$ booster refers to R-Frame followed by Mix-up operation for data generation; 
    \edit{Baseline Ensemble refers to ConvNet variant of \cite{guan2017}, and it is a special case of our ConvBoost when without boosters.}
    }
    \label{extension_table}
    \small
    \begin{tabular}{l|l|l|l|c|c|c}
    \toprule
    \multicolumn{1}{c|}{\multirow{2}[1]{*}{Methods}} & \multicolumn{3}{c|}{Layers (with boosters)} & \multicolumn{3}{c}{Backbones} \bigstrut\\
    & Sampling Layer & Data Aug. Layer & Resilient Layer & CNN & ConvLSTM & Att. Model \bigstrut\\
    \midrule
    Baseline Ensemble & \multicolumn{1}{c|}{-} & \multicolumn{1}{c|}{-} & \multicolumn{1}{c|}{-} & 65.61 $\pm$ 1.23 & 65.50 $\pm$ 1.11 & 67.58 $\pm$ 1.23 \bigstrut\\
    \hline
   \edit{ConvBoost Ensemble} &  \multicolumn{1}{c|}{R-Frame} &  \multicolumn{1}{c|}{Mix-up} & \multicolumn{1}{c|}{-}  & \edit{69.86 $\pm$ 0.88} & \edit{70.27 $\pm$ 0.94} & \edit{71.86 $\pm$ 1.11} \bigstrut\\
    \edit{ConvBoost Ensemble} & \multicolumn{1}{c|}{-} &  \multicolumn{1}{c|}{Mix-up}&   \multicolumn{1}{c|}{C-Drop} & \edit{66.65 $\pm$ 1.18} & \edit{67.53 $\pm$ 1.10} & \edit{71.28 $\pm$ 1.01} \bigstrut\\
    \edit{ConvBoost Ensemble} &  \multicolumn{1}{c|}{R-Frame} & \multicolumn{1}{c|}{-} & \multicolumn{1}{c|}{C-Drop}  & \edit{69.46 $\pm$ 0.94} & \edit{69.58 $\pm$ 0.74} & \edit{70.43 $\pm$ 0.98} \bigstrut\\
    \midrule
    \edit{ConvBoost Ensemble} &  \multicolumn{1}{c|}{R-Frame}  &  \multicolumn{1}{c|}{Mix-up} &  \multicolumn{1}{c|}{C-Drop}  & 70.53 $\pm$ 0.81 & 71.24 $\pm$ 0.96 & 72.81 $\pm$ 0.76 \bigstrut\\
    \midrule
    \multicolumn{1}{c|}{Exp. 1} &  \multicolumn{1}{c|}{R-Frame}  & Mix-up;   R-Frame$^{*}$ &  \multicolumn{1}{c|}{C-Drop}  & \textbf{71.21 $\pm$ 1.03} & 71.39 $\pm$ 1.05 & 73.01 $\pm$ 0.88 \bigstrut\\
    \multicolumn{1}{c|}{Exp. 2}& \multicolumn{1}{c|}{R-Frame} &  \multicolumn{1}{c|}{Mix-up} & C-Drop;   Scaling & 70.71 $\pm$ 0.88 & \textbf{71.66 $\pm$ 0.70} & \textbf{73.07 $\pm$ 1.12} \bigstrut\\
    \midrule
    \end{tabular}%
\end{table}%

\subsubsection{Compression \edit{in ConvBoost Ensemble }}
\edit{When in Ensemble mode, it is necessary to}
explore ensemble compression for efficient applications.
Here we simply average the models' parameters, which can reduce the model size by $95\%$ (i.e., from 20 base classifiers to 1 classifier).
Due to the lack of diversity in the ensemble (as shown in \autoref{cs_qs_result}), the model parameters may be regarded as a multivariate Gaussian distribution with \edit{low variance} in each dimension, and in this case the corresponding mean model serves as light-weight representation for the whole distribution (the ensemble).

It is worth pointing out this idea is very similar to the recent proposed model soup \cite{wortsman2022model}, where a number of pre-trained models fine-tuned based on different hyper-parameters were averaged for prediction. 
To the best of our knowledge, there is no feasible pre-trained models in the HAR community that generalize well, and here we employed the parameter averaging idea \edit{for ConvBoost Ensemble compression}.

\begin{figure}[t]
  \centering
  \includegraphics[width=\textwidth]{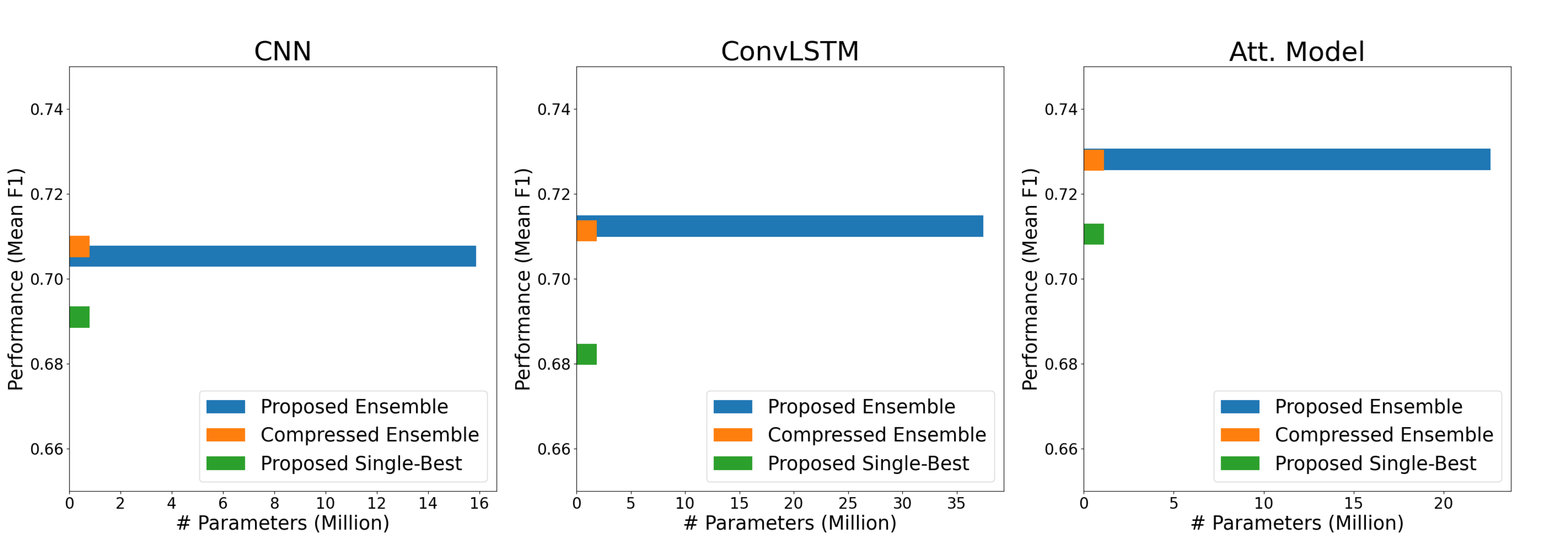}
  \caption{Model Comparison in terms of Performance (i.e., $\bar{F}_1$ in y axis)  and Model size (i.e, number of parameters in x axis) on OPP dataset; \edit{Proposed Ensemble (or Single-Best) refer to the proposed ConvBoost framework in Ensemble mode (or Single-Best mode). 
  }
  }
  \label{compression}
  \vspace*{-1em}
\end{figure}

On the OPP dataset, we compare the results of \edit{the proposed ConvBoost Ensemble, the compressed counterpart (via parameter averaging), and the proposed ConvBoost Single-Best} in terms of both performance and model size, as shown in \autoref{compression}.  
We can see via the simple model averaging, the compressed model has comparable performance with \edit{the ensemble counterpart, but} with only $5\%$ of the parameters.
On the other hand, although ConvBoost Single-Best is also a strong light-weight method (cf.\  \autoref{result_all}), the proposed simple compressed ensemble outperforms Single-Best at the same model size irrespective of backbones, suggesting 
\edit{it is a practical solution for light-weight applications.}
In the future, we will explore more advanced model compression approaches for higher performance in both effectiveness and efficiency.

\subsubsection{Potential Applications and Future Work}
In our 3-layer ConvBoost framework, we design and employ R-Frame, mix-up, and C-Drop boosters for ConvNet-based HAR. 
Out of them, R-Frame and mix-up are generic, while C-Drop is domain-specific that is designed to simulate the problematic sensor data. 
We observed the performance of our ConvBoost is very competitive even only with the generic boosters ( i.e., R-Frame and mix-up boosters here), and this indicates the great potentials of our framework on the general time-series analysis applications.
For various application, under the ConvBoost framework one can focus on designing domain-specific boosters for performance enhancement. 

Although we demonstrated our ConvBoost framework can exploit the potentials of (limited) labeled sequential sensor data, all the boosters were developed to facilitate supervised learning. 
In the future, we will explore how to develop boosters for unsupervised learning/self-supervised learning schemes to deal with vast amount of unlabeled data collected in unconstrained environments. 
 


%% file: sections/conclusion.tex
We proposed the ConvNet-Boosting (ConvBoost) Framework, which includes three conceptual layers--Sampling Layer, Data Augmentation Layer, and Resilient Layer--aiming at boosting the performance of ConvNet-based HAR models.
Within the three conceptual layers, we designed three boosters--R-Frame, Mix-up, and C-Drop--which can generate per-training-epoch additional informative training frames via dense sampling, synthesizing, and simulating operations.
With the epoch-wise generated data, the base classifiers trained from each epoch can be effectively strengthened, yielding significant performance gain in both Single-Best and Ensemble modes irrespective of ConvNet types for HAR.

Through our ConvNet, we also re-interpreted  previous work on epoch-wise bagging schemes \cite{guan2017}, and developed its ConvNet variant.
Through our extensive experiments we found--for the Ensemble mode--that the performance gains were mainly from the strengthened individual (epoch-wise) classifiers, rather than ensemble diversity.
In this case, our 3-layer ConvBoost can generate diverse complementary training examples from different perspectives, which substantially enhances the base learners, yielding improved performance in both Ensemble mode and Single-Best mode. 
We also explored two extensions by adding more boosters and applying compression to ConvBoost Ensemble, and very encouraging initial results were achieved, indicating it is an extensible and flexible solution for various HAR tasks.

%% file: sections/appendix.tex

\section{Q\_statistics (QS) for Diversity Measurement} 

Q\_statistics (QS) \cite{udny1900association} is a popular metric to measure the ensemble diversity.
Specifically, it computes the pairwise relationship between any two predictions (from base learners) in the ensemble. 


Given a query data, assuming $\hat{y}_i$ and $\hat{y}_k$ are two predictions from two base classifiers (corresponding to the $i^{th}$ and the $k^{th}$ classifiers in the ensemble), we set $\hat{y}_i=1$ (or $\hat{y}_k=1$) for correct predictions and $\hat{y}_i=0$ (or $\hat{y}_k=0$) for incorrect predictions.  
Given a dataset,  
$QS(i,k)$ for the $i^{th}$ and the $k^{th}$ classifiers can be defined as 
\begin{equation}
{QS(i, k)}=\frac{{N^{11}}{N^{00}}-{N^{01}}{N^{10}}}{{N^{11}}{N^{00}}+{N^{01}}{N^{10}}},
\end{equation}
where 
$N^{ab}$ is defined as the occurrence of the following prediction scenarios of both classifiers, i.e.,
\begin{equation}
N^{ab} =  \left\{ 
  \begin{array}{ c l }
    N^{11} & \quad \textrm{if } \hat{y}_{i} = 1, \hat{y}_{k} = 1 \\
    N^{10} & \quad \textrm{if } \hat{y}_{i} = 1, \hat{y}_{k} = 0 \\
    N^{01} & \quad \textrm{if } \hat{y}_{i} = 0, \hat{y}_{k} = 1 \\
    N^{00} & \quad \textrm{if } \hat{y}_{i} = 0, \hat{y}_{k} = 0 \\
  \end{array}
\right.
\end{equation}
We can further extend it to an ensemble with $M$ base learners, with QS defined as: 
\begin{equation}
{QS}=\frac{1}{M^2}\sum_{i=1}^{M}\sum_{k=1}^{M}\frac{{N^{11}}{N^{00}}-{N^{01}}{N^{10}}}{{N^{11}}{N^{00}}+{N^{01}}{N^{10}}}
\end{equation}
to measure the pairwise relationship of base learners' predictions in the ensemble.

\section{Additional Results}
\subsection{Confusion Matrices on Three datasets}
\begin{figure}[htbp]
  \includegraphics[width=0.45\textwidth, trim={0cm 0cm 2.5cm 0cm},clip]{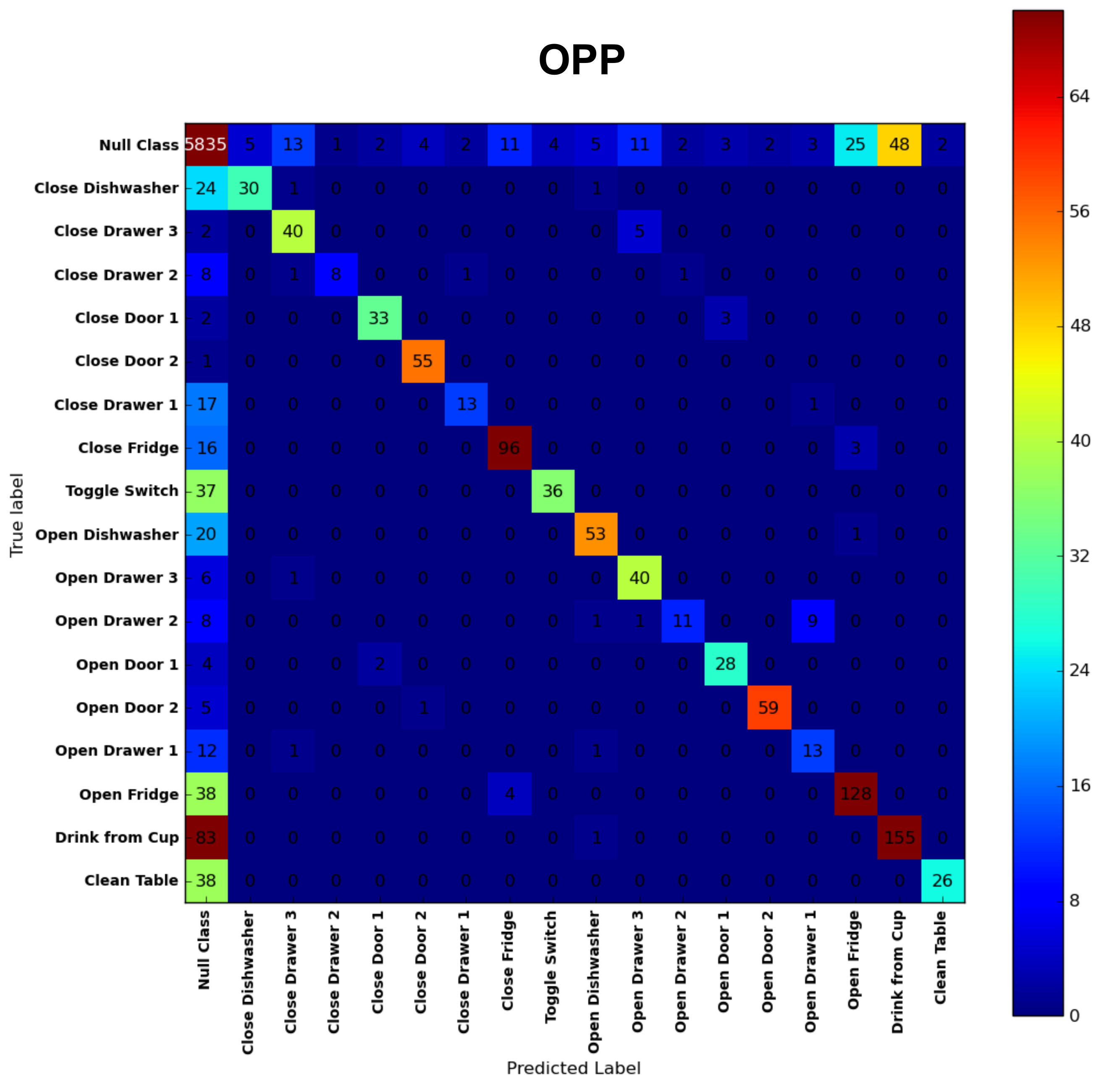}
  \includegraphics[width=0.45\textwidth, trim={1cm 1cm 2.5cm 0cm},clip]{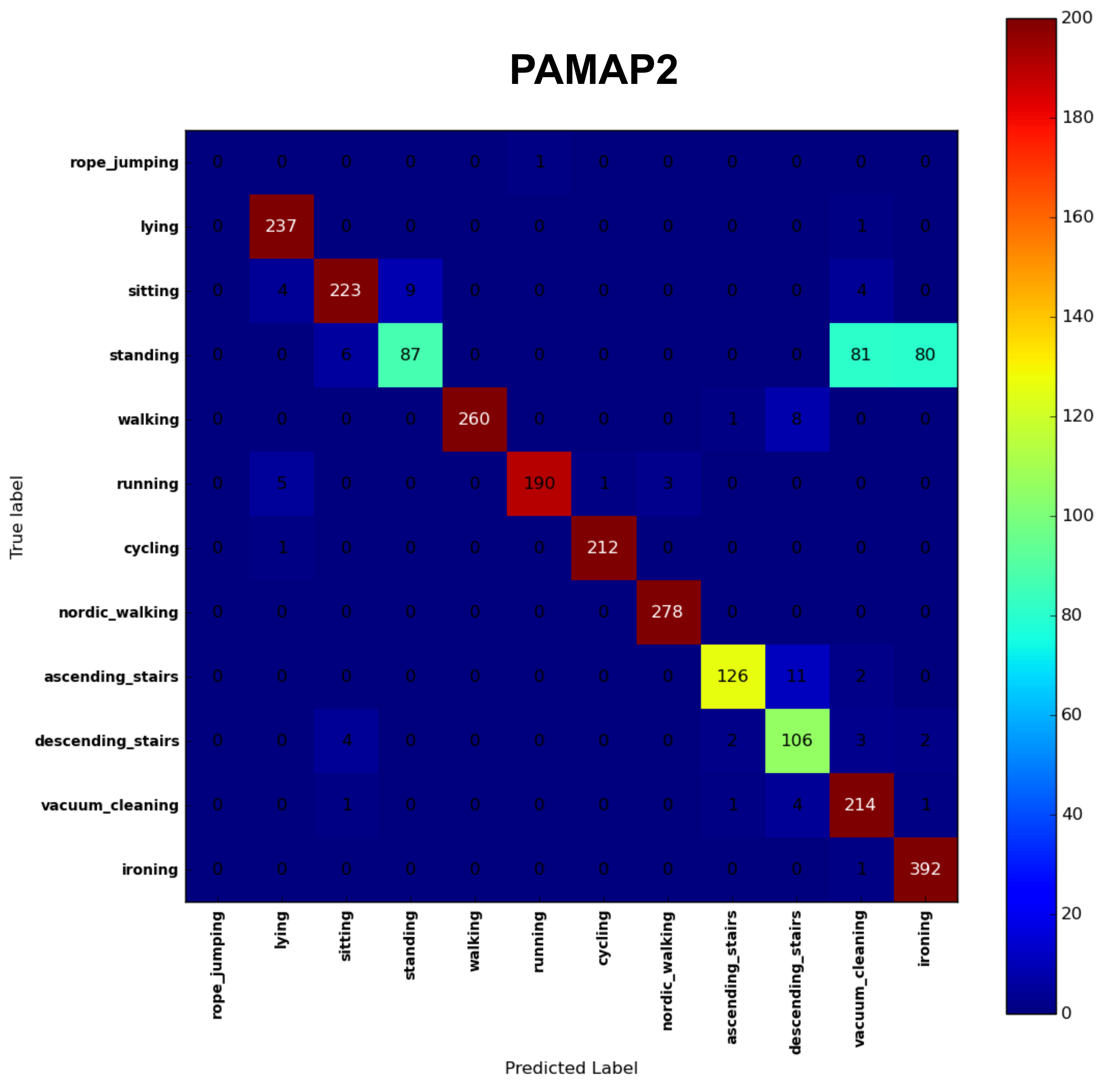} 
  \includegraphics[width=0.45\textwidth, trim={0cm 0cm 1cm 0cm},clip]{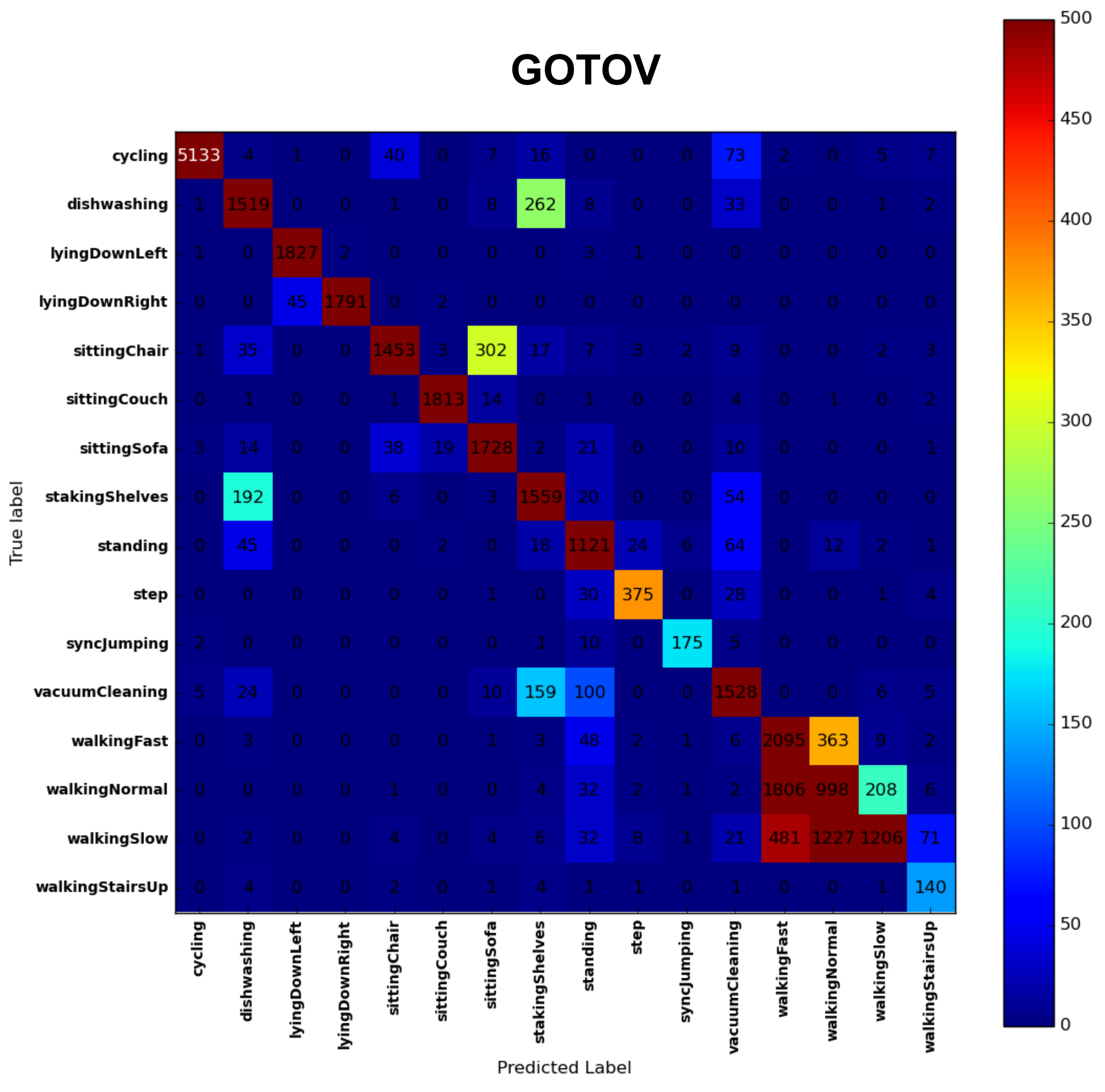} 
\caption{Confusion matrices of the Proposed ConvBoost Ensemble (with Attention model as backbone) on three datasets. The results are the averaged values of 20 repetitions. Best viewed in color.
}
  \label{confusion_matrix_att_model}
\end{figure}

Based on the proposed ConvBoost Ensemble (with Attention Model backbone), in \autoref{confusion_matrix_att_model} we report the confusion matrices of the three datasets.
Although ConvBoost can boost the performance of ConvNet-based models via generating diverse training frames, the HAR challenges still exist in various scenarios. 
For example, on GOTOV dataset, our model cannot easily distinguish the activity classes "Walking slow", "Walking fast","Walking normal", which has much higher inter-class similarity. 
On OPP dataset, the error patterns are mainly from the "Null class", which is the majority class taking up to $75\%$ of the total data, yielding lots of false negative predictions.

\subsection{Results on PAMAP2 Dataset in Leave-One-Subject-Out Setting}
\begin{table}[htbp]
\caption{
Performance comparison (i.e., $\bar{F}_{1}$ in $\%$,) between baseline ensemble (i.e., ConvNet variant of \cite{guan2017}) and our proposed ConvBoost ensemble on the PAMAP2 dataset in leave-one-subject-out setting;  
In this setting, subject 9 was excluded due to the lack of activity classes.
We ran each model 20 repetitions with the corresponding mean and standard deviation values reported here. 
}

\begin{tabular}{@{}ccccccc@{}}
\toprule
\multirow{2}{*}{Subject} & \multicolumn{3}{c}{Baseline Ensemble} & \multicolumn{3}{c}{ConvBoost Ensemble  (Proposed)} \\ \cmidrule(l){2-7} 
& CNN               & ConvLSTM          & Att. Model       & CNN              & ConvLSTM         & Att. Model       \\ \cmidrule(r){1-7}                  
1& 71.77 $\pm$ 2.33      & 69.44 $\pm$ 2.03      & 75.14 $\pm$ 4.52     & 77.19 $\pm$ 1.56      & 75.63 $\pm$ 0.94     & 79.09 $\pm$ 1.25     \\
2                        & 83.80 $\pm$ 4.29      & 77.66 $\pm$ 3.90      & 86.04 $\pm$ 2.83     & 88.96 $\pm$ 3.32      & 90.77 $\pm$ 3.73     & 93.77 $\pm$ 1.45     \\
3                        & 68.34 $\pm$ 6.58      & 69.64 $\pm$ 6.02      & 75.09 $\pm$ 7.71     & 72.40 $\pm$ 5.62      & 69.56 $\pm$ 6.15     & 77.55 $\pm$ 6.16     \\
4                        & 81.17 $\pm$ 4.27      & 79.40 $\pm$ 4.00      & 85.15 $\pm$ 4.67     & 92.31 $\pm$ 2.79      & 88.81 $\pm$ 4.44     & 93.16 $\pm$ 4.44     \\
5                        & 88.27 $\pm$ 2.23      & 82.84 $\pm$ 2.77      & 93.02 $\pm$ 0.53     & 92.60 $\pm$ 0.34      & 92.75 $\pm$ 0.32     & 93.15 $\pm$ 0.34     \\
6                        & 86.82 $\pm$ 3.25      & 83.60 $\pm$ 2.35      & 88.99 $\pm$ 3.31     & 89.83 $\pm$ 2.80      & 90.09 $\pm$ 3.05     & 90.38 $\pm$ 3.98     \\
7                        & 92.51 $\pm$ 4.29      & 95.07 $\pm$ 2.65      & 95.82 $\pm$ 2.04     & 96.08 $\pm$ 2.66      & 93.04 $\pm$ 4.36     & 95.21 $\pm$ 3.59     \\
8                        & 38.23 $\pm$ 4.73      & 41.42 $\pm$ 3.15      & 70.56 $\pm$ 7.48     & 47.71 $\pm$ 3.65      & 45.38 $\pm$ 2.91     & 70.09 $\pm$ 5.59     \\ \cmidrule(l){1-7} 
Avg.                     & 76.36            & 74.88             & 83.73            & 82.13             & 80.75            & 86.54            \\ \bottomrule
\end{tabular}
\label{loso_pamap2}
\end{table}